%% file: main_arxiv.tex
\documentclass[]{beingbeyond}
\usepackage{enumitem}
\usepackage[utf8]{inputenc} % allow utf-8 input
\usepackage[T1]{fontenc}    % use 8-bit T1 fonts
\usepackage{hyperref}       % hyperlinks
\usepackage{url}            % simple URL typesetting
\usepackage{array}          % for advanced table column types
\usepackage{booktabs}       % professional-quality tables
\usepackage{amsfonts}       % blackboard math symbols
\usepackage{nicefrac}       % compact symbols for 1/2, etc.
\usepackage{microtype}      % microtypography
\usepackage{xcolor}         % colors
\usepackage{xspace}
\usepackage{bm}
\usepackage{bbm}
\usepackage{bbding}
\usepackage{tabularx}
\usepackage{textcomp}
\usepackage{amssymb}
\usepackage{enumitem}
\usepackage{amsmath}
\usepackage{mathtools}
\usepackage{amsthm}
\usepackage{multirow}
\usepackage{makecell}
\usepackage{color}
\usepackage{colortbl}
\usepackage{adjustbox}
\usepackage{caption}
\usepackage{graphicx}
\usepackage{wrapfig}
\usepackage{array}
\usepackage{multicol}
\definecolor{myyellow}{RGB}{255,192,0}
\definecolor{mygreen}{RGB}{107,170,64}
\definecolor{mywrite}{RGB}{255,227,132}
\title{Joint-Aligned Latent Action: Towards Scalable\\ VLA Pretraining in the Wild}

\author{{\bfseries 
Hao Luo$^{1,3}$ \quad 
Ye Wang$^{2,3}$ \quad
Wanpeng Zhang$^{1,3}$ \quad
Haoqi Yuan$^{1,3}$ \quad
\\
Yicheng Feng$^{1,3}$ \quad
Haiweng Xu$^{1,3}$ \quad
Sipeng Zheng$^{3}$  \quad
Zongqing Lu$^{1,3,\dagger}$
}}

\affiliation{{$^{1}$Peking University \quad $^{2}$Renmin University of China \quad $^{3}$BeingBeyond}}

\webpage{\url{https://research.beingbeyond.com/jala}}

\input{sections_arxiv/00_abstract}

\checkdata[Date]{Feb 25, 2026}

\definecolor{BlockC}{gray}{0.98}  
\definecolor{BlockA}{RGB}{191,211,230}
\definecolor{BlockB}{RGB}{199,233,192}

\begingroup
\footnotetext[2]{Correspondence to Zongqing Lu $<$lu@beingbeyond.com$>$.}
\endgroup

\begin{document}

\maketitle

\input{sections_arxiv/01_introduction_fixed}

\input{sections_arxiv/02_related_work_condensed}
\input{sections_arxiv/03_formulation_dataset}
\input{sections_arxiv/04_methodology_new}
\input{sections_arxiv/05_dataset}

\input{sections_arxiv/06_experiments}

\input{sections_arxiv/07_conclusion}

\clearpage

\bibliographystyle{unsrt}
\bibliography{bibliography/references}

%\clearpage

%\beginappendix

%\section{appendix section}

%appendix

\clearpage
% No appendix in arXiv version; all content is merged into main sections.

\end{document}

%% file: sections_arxiv/00_abstract.tex
\abstract{
Despite progress, Vision-Language-Action models (VLAs) are limited by a scarcity of large-scale, diverse robot data. 
While human manipulation videos offer a rich alternative, existing methods are forced to choose between small, precisely-labeled datasets and vast in-the-wild footage with unreliable hand tracking labels.
We present \textbf{JALA}, a pretraining framework that learns \textbf{J}ointly-\textbf{A}ligned \textbf{L}atent \textbf{A}ctions.
JALA bypasses full visual dynamic reconstruction, instead learns a predictive action embedding aligned with both inverse dynamics and real actions.
This yields a transition-aware, behavior-centric latent space for learning from heterogeneous human data.
We scale this approach with \textbf{UniHand-Mix}, a 7.5M video corpus (>2,000 hours) blending laboratory and in-the-wild footage.
Experiments demonstrate that JALA generates more realistic hand motions in both controlled and unconstrained scenarios, significantly improving downstream robot manipulation performance in both simulation and real-world tasks.
These results indicate that jointly-aligned latent actions offer a scalable pathway for VLA pretraining from human data.}

%% file: sections_arxiv/01_introduction_fixed.tex
\section{Introduction}
\label{sec:introduction}

Researchers have long sought foundation models enabling robots to perform diverse tasks.
Recent Vision-Language-Action models (VLAs)~\citep{zitkovich2023rt,black2024pi_0,bjorck2025gr00tn1} show promise by adapting Large Multimodal Models (LMMs) with robotics datasets~\citep{o2024open,khazatsky2024droid}.
However, robotic data remains orders of magnitude smaller and less diverse than datasets enabling breakthroughs in vision and language~\citep{touvron2023llama, li2024llava}.
This scarcity, combined with cross-embodiment heterogeneity, creates a major obstacle for scaling VLA pretraining~\citep{kim2025openvla, team2024octo}. 

To overcome this data bottleneck, recent work~\citep{beingbeyond2025beingh0,yang2025egovla} has turned to human demonstration videos by interpreting hand motion as a rich, cost-effecient signals for training, which allows VLAs to acquire skills beyond robot datasets.
%Interpreting hand motion as a form of action allows VLAs to acquire skills beyond robot-only datasets.
However, a quality-variety trade-off persists: lab-collected datasets~\citep{fan2023arctic,chao2021dexycb,egodex, liu2022hoi4d} provide precise 3D hand tracking but are limited to controlled tabletop scenarios, while in-the-wild videos~\citep{grauman2022ego4d,perrett2025hd,damen2022rescaling} offer immense diversity and natural behaviors but lack reliable action labels. 
This leads to our core question: \textit{how can we effectively combine these complementary human data sources to scale VLA pretraining?}

\begin{figure*}[t]
    \centering
    \includegraphics[width=0.95\linewidth]{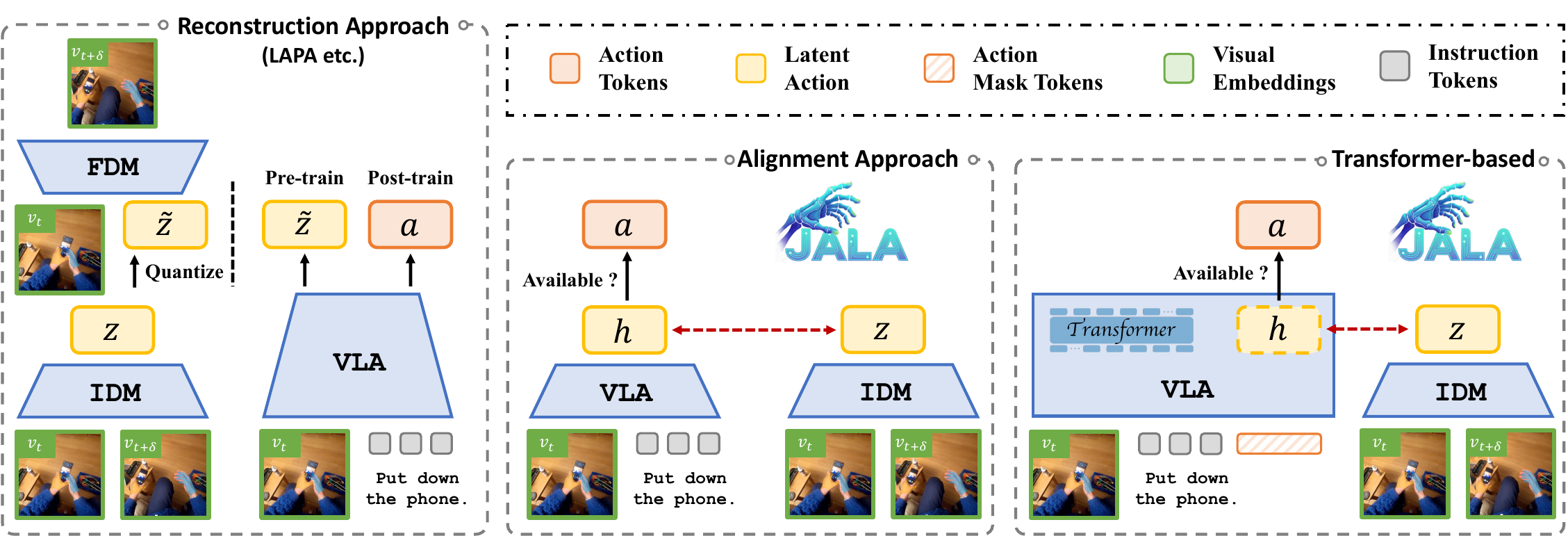}
    % \vspace{-10pt}
    \caption{\textbf{Comparison of VLA's latent action paradigms with human videos.} 
    \textbf{(left)} Prior reconstruction-based methods like LAPA~\citep{ye2024lapa} rely on multi-stage pipelines extracting latent actions via dynamics reconstruction as pseudo-labels. 
    \textbf{(middle)} Our JALA introduces predictive embeddings aligned with latent actions. 
    \textbf{(right)} Transformer-based JALA implementation where intermediate hidden states serve as the predictive embeddings to align with latent actions, while output tokens use available action labels as supervision.}
    \label{fig:concise_abl}
    \vspace{-10pt}
\end{figure*}

Although in-the-wild videos lack action labels, their motion \textbf{dynamics} reflect the visual changes during interactions. 
Prior approaches capture this indirectly via auxiliary visual representation learning~\citep{nair2023r3m,radosavovic2023real} or world models~\citep{wu2024gr1,cheang2024gr2}, but their connection to action is weak.
To extract more direct action information, recent works~\citep{ye2024lapa,video2skill2024,chen2024igor} exploit \textbf{inverse dynamics}, which maps visual transitions to underlying actions, producing \textbf{latent actions} that form a \textbf{latent action space}.
Typically, an inverse dynamics model (IDM) infers latent actions from current and future frames, while a forward dynamics model (FDM) reconstructs future frames from latent actions and current frames, constraining latent actions to encode action-relevant information (Fig.~\ref{fig:concise_abl}, left).
However, for fine-grained human manipulation, this paradigm is inefficient because the quality of latent actions depends heavily on the FDM's ability to model future frames. 
The subtlety and variability of hand motions make learning an accurate FDM challenging, often introducing noise that degrades the latent action quality rather than stable grounding.

In this paper, we propose a novel perspective on human actions for VLA pretraining, inspired by how humans learn manipulation through transferable action patterns rather than memorizing every visual detail.
Our framework, \textbf{JALA} (\textbf{J}oint-\textbf{A}ligned \textbf{L}atent \textbf{A}ctions), replaces reconstruction-driven pipelines (Fig.~\ref{fig:concise_abl}, left) with \textbf{joint alignment} between a \emph{predictive embedding} $h$ produced from the VLA context and the \emph{latent action} $z$ derived from IDM. 
This shapes a latent action space that is predictable from context while remaining informative of inverse dynamics, without reconstructing pixels.
When available, ground-truth action $a$ is also predicted from $h$, grounding the shared space with explicit control semantics through predictive embeddings. 
%Since joint alignment is applicable 
Regardless of action-label availability, JALA constructs a unified latent action space that supports learning from both lab-annotated and in-the-wild videos.

To realize this design, we instantiate JALA on a Transformer-based VLA.
As shown in Fig.~\ref{fig:concise_abl} (right), the VLA predicts action tokens from action mask tokens while aligning intermediate hidden state with latent actions from IDM --- these hidden states serve as predictive embeddings connecting real actions and latent actions. 
When we adapt the pretrained VLA for robot tasks, we feed these predictive embeddings into a flow-matching head~\citep{lipman2022flow}, efficiently transferring the pretrained latent action space to robot tasks.

We scale pretraining with \textbf{UniHand-Mix}, a hybrid 7.5M-sample corpus combining 5M lab-recorded videos with precise hand tracking and 2.5M in-the-wild manipulation clips without it. 
Following~\cite{beingbeyond2025beingh0}, we process 1,000 hours of lab videos to create instruction-tuning samples with explicit motion labels, then extend to unconstrained settings by filtering video clips from another 1,123 hours of Ego4D~\citep{grauman2022ego4d}. 
Our JALA is evaluated on hand motion generation and robot manipulation tasks. 
Results show JALA excels at generating realistic hand motions in wild scenarios while maintaining laboratory performance.
It outperforms similar-size VLAs on Libero~\citep{liu2023libero} and RoboCasa~\citep{robocasa2024}, remains competitive with larger models, and achieves superior real-world performance, especially in out-of-distribution tasks.

Our key contributions are: 
(1) \textbf{Novel latent action paradigm:}
We introduce joint-aligned latent actions, enabling VLAs to learn from both labeled and unlabeled human videos at an unprecedented scale.
(2) \textbf{Large-scale hybrid dataset:}
We create UniHand-Mix, extending UniHand with 2.5M in-the-wild human manipulation samples, providing greater diversity for VLA pretraining.
(3) \textbf{Strong empirical results:}
We demonstrate a VLA that substantially improves generalization in hand motion generation with SoTA performance among similar-size models on robot tasks.

%% file: sections_arxiv/02_related_work_condensed.tex
\section{Related Work}
\label{sec:related}

\paragraph{Vision-Language-Action Model Pretraining.} 
There has been growing interest in building VLAs by adapting pretrained vision-language architectures for robotics. 
Using large-scale robot datasets~\citep{open2024open,khazatsky2024droid,bu2025agibot}, researchers have taken two main approaches: one convert robot actions into discrete tokens and trains like autoregressive LLMs~\citep{brohan2022rt1,zitkovich2023rt,o2024open,kim2025openvla,pertsch2025fast,zhong2025survey}, while another uses diffusion- or flow-matching methods to generate continuous actions~\citep{team2024octo,black2024pi_0,intelligence2025pi_,bjorck2025gr00tn1,driess2025knowledge,liu2024rdt,wen2025dexvla,shukor2025smolvla}. 
In our work, we use tokenization during pretraining to learn general action patterns, then switch to flow-matching for precise robot control during deployment.
However, robot data alone is limited, so researchers have turned to human videos as a more scalable resource. 
Some works learn indirectly from human videos by extracting generic visual representations or building world-models ~\citep{bjorck2025gr00tn1,radosavovic2023real,nair2023r3m,wu2024gr1,cheang2024gr2}, or creating latent action spaces~\citep{ye2024lapa,univla2024,chen2024igor}.
While promising, these approaches struggle to connect abstract representations to actual physical actions. 
Other approaches use lab-collected human data with precise hand tracking~\citep{egodex, banerjee2025hot3d, liu2022hoi4d, chao2021dexycb} to explicitly model human hand movements~\citep{beingbeyond2025beingh0,yang2025egovla,singh2025deep}.
This offers stronger physical grounding but requires expensive data collection and limits task diversity. 
Our work bridges these two approaches. 
We introduce a new method for constructing latent actions by aligning predictive embeddings with motion dynamics, allowing us to combine the scalability of learning from unlabeled videos with the precision of explicitly annotated trajectories.
Our design reduces annotation cost while maintaining physical grounding, enabling VLA pretraining to scale more effectively across heterogeneous human datasets.

\paragraph{Learning from Videos.}
Our work also connects to research on \textit{learning from videos} (LfV)~\citep{yang2015robot,seo2022reinforcement}, where studies extract behavioral knowledge from unlabeled videos using various techniques: masked auto-encoding which predicts missing parts of videos~\citep{radosavovic2023real,xiao2022masked}, temporal contrastive learning which learn to distinguish different video clips~\citep{li2024auxiliary,nair2023r3m}, video prediction which forecasts future frames~\citep{seo2022reinforcement,luo2024pre}, or inverse dynamics modeling which infers actions from visual changes~\citep{baker2022video,schmeckpeper2021reinforcement,ye2022become}. 
For robot control specifically, some works exploit carefully matched human–robot pairs to enable imitation policies~\citep{kareer2024egomimic, singh2024hand,luo2025learning, zheng2025flare}, but this approach only scales to specific task sets. 
When using unconstrained human videos with robot actions~\citep{univla2024, ye2024lapa}, effectiveness is limited due to unclear alignment between human and robot behaviors.
Our approach differs by leveraging human hand tracking as a rich source of physical alignment, with the human hand serving as the most extensively annotated manipulator available.
We combine these precise annotations with diverse in-the-wild videos within a unified pretraining framework, achieving both physical grounding and task diversity.

%% file: sections_arxiv/03_formulation_dataset.tex
\section{Preliminaries}
\label{sec:hand_motion_modeling}

%We introduce preliminaries for JALA, detailing human hand movement modeling during VLA pretraining and our hybrid data formulation with optional pose annotations.

Following Being-H0~\cite{beingbeyond2025beingh0}, we model human hand movements for VLA pretraining using MANO parameters~\citep{romero2017mano}.
Each data sample contains a video $v = \{v_1, v_2, \ldots, v_T\}$, a text instruction $x$, and a hand pose sequence $\mathcal{M} = \{m_1, m_2, \ldots, m_T\}$, where each pose $m_t$ is defined by MANO parameters $(\theta_t, \mathbf{r}_t, \tau_t, \beta_t)$ for relative joint angles, global wrist rotation, wrist translation, and hand shape (the static shape parameter $\beta_t$ is excluded). 
For pretraining, we discretize the pose sequence into fixed-length chunks, tokenizing each into $K$ motion tokens via GRQ~\citep{yang2023hifi}. 
The input training sequence is: $[x, v_1, A_1, A_2, \ldots, A_N]$, where $v_1$ is the first frame and $A_i$ denotes the $i$-th tokenized motion chunk with length $T / N$.
%With chunk length $\delta$ and $N = T / \delta$ chunks, the training sequence becomes: $[x, v_1, A_1, A_2, \ldots, A_N]$, where $v_1$ is the first video frame and $A_i$ denotes the $i$-th tokenized motion chunk.
This format aligns with the VLA pretraining paradigm, allowing to generate hand motions based on visual-text inputs.
Our VLA $f_\Theta$ is trained to maximize motion token likelihood given the previous context:
\begin{equation}
\max_\Theta \sum_{i=1}^N \log p(A_i \mid A_{<i}, v_1, x; \Theta).
\end{equation}
We also include auxiliary training tasks (e.g., motion description generation, motion continuation), which help the model learn richer token semantics before adapting to robot tasks while keeping the formulation of generating motions chunk by chunk.
Given limited availability of hand tracking data, we use both annotated and in-the-wild videos.
Here, we define:
(1) Annotated dataset $\mathcal{D}_A = \{(x^{(i)}, v^{(i)}, \mathcal{M}^{(i)})\}_{i=1}^{N_A}$ with instructions, videos, and hand poses. 
(2) Unannotated dataset $\mathcal{D}_U = \{(x^{(j)}, v^{(j)})\}_{j=1}^{N_U}$ with only instructions and videos.
Our goal is to jointly train on the hybrid dataset $\mathcal{D} = \mathcal{D}_A \cup \mathcal{D}_U$, leveraging both the precision of annotated data and the diversity of in-the-wild videos.

%% file: sections_arxiv/04_methodology_new.tex
\section{Methodology}
\label{sec:methodology}
\begin{figure*}
\centering
\includegraphics[width=\linewidth]{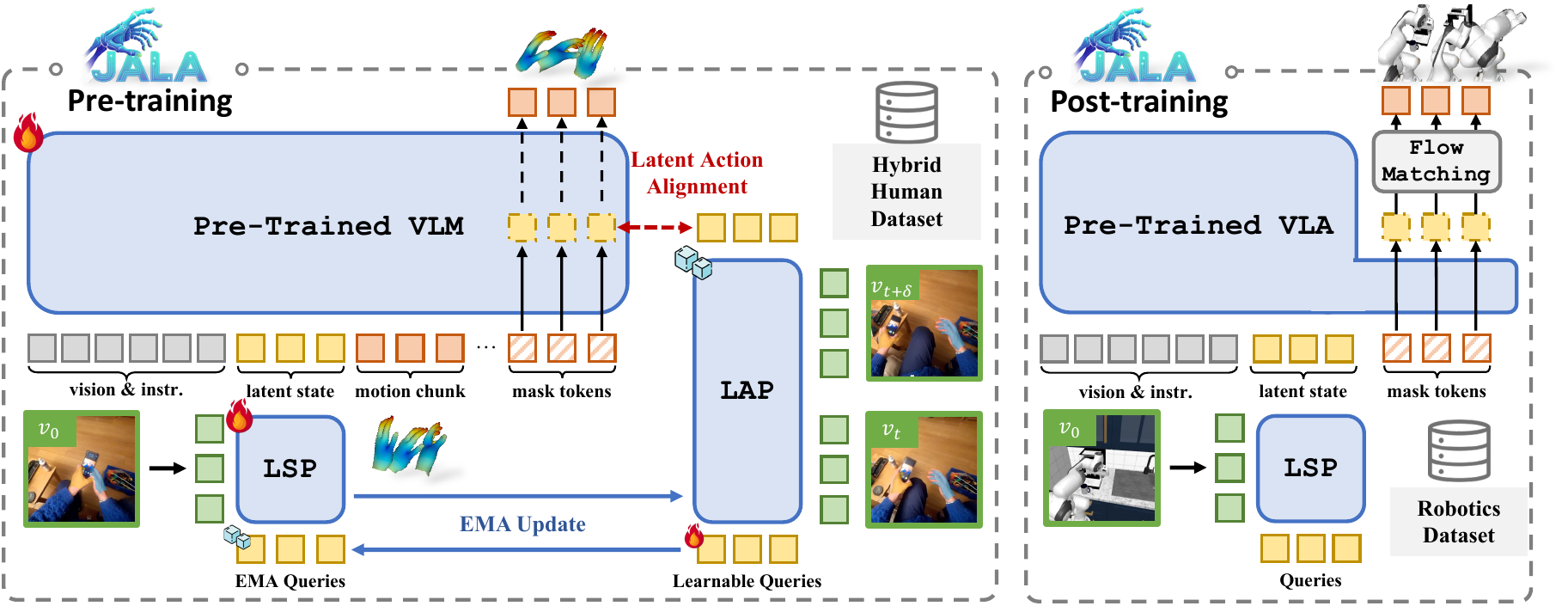}
\caption{\textbf{The JALA framework}. 
\textbf{Pre-training (left)}:
Hidden states of masked motion chunks serve as predictive embeddings to align with latent actions from boundary frames.
The Latent Action Perceiver (LAP) maps boundary frames to latent action space, providing supervision without action labels. 
A parameter-shared Latent State Perceiver (LSP) injects initial frame context, with LAP and LSP linked via decoupled EMA update for stability. 
\textbf{Post-training (right)}: 
The predictive embeddings are fed into a flow-matching head for robot task transfer.}
\label{fig:JALA_arch}
\end{figure*}

As illustrated in Fig.~\ref{fig:JALA_arch}, JALA builds on a Transformer-based vision-language model that processes visual tokens, instruction tokens, and motion tokens together. In JALA, we use the intermediate hidden states as \textit{predictive embeddings} to conduct \textbf{joint alignment} with \textit{latent actions} inferred from inverse dynamics. Accompanied with optional action supervision, the joint alignment derives a unified latent action space predictable from the VLA context while encoding both visual dynamics and action semantics across the hybrid human video dataset.

\subsection{Joint Alignment with Latent Actions}
\label{sec:pred_align}
 
For each motion token $a_{i,k}$ in a tokenized chunk $A_i$, we extract its hidden state from a preselected attention layer as the predictive embedding, denoted as $h_{i,k} \in \mathbb{R}^d$. The predictive embedding is the core of joint alignment and is shaped by two complementary signals: 
(1) supervision from motion labels, and 
(2) alignment with latent actions.
Below, we detail joint alignment with these two signals, respectively.

\paragraph{Masked Chunk Prediction (MCP).}
To capture motion information based on hand tracking labels, we introduce MCP, a chunk-level masked token modeling objective similar to GR-1~\citep{wu2024gr1}. 
During pretraining, we replace all motion tokens in a chunk with \texttt{[MASK]} placeholder and use bidirectional attention within the chunk.
This allows the model to understand relationships between motions within a chunk.
The VLA learns by predicting the original tokens:
\begin{equation}
\mathcal{L}_{\text{MCP}}
= - \sum_{i=1}^N \sum_{k=1}^K 
\log p_\Theta \big(a_{i,k} \mid A_{<i}, v, x \big).
\end{equation}
Although we sum over individual tokens, the bidirectional attention ensures that all motion tokens $a_{i,k}$ in a chunk $A_i$ are modeled jointly. Thus, the corresponding hidden states $h_{i,k}$ carry the chunk-level movement patterns via a predictive embedding in the VLA context.

\paragraph{Latent Action Perceiver (LAP).}
To jointly align predictive embeddings with visual dynamics, we introduce LAP, a Perceiver~\cite{jaegle2021perceiver,shridhar2023perceiver} module, as an inverse dynamics model. 
For each chunk, LAP takes the start and end frames $(v_t, v_{t+\delta})$ and produces $K$ latent action vectors $\{z_{i,1}, \ldots, z_{i,K}\}$ using a fixed set of learnable query vectors.  
These latent actions capture the dynamics of chunk-level transition. 
Instead of reconstruction on full video frames, we jointly align the predictive embeddings $h_{i,k}$ with the latent actions $z_{i,k}$:
\begin{equation}
\mathcal{L}_{\text{Align}}
= \sum_{i=1}^N \sum_{k=1}^K \| h_{i,k} - z_{i,k} \|_1.
\end{equation}
Individually, MCP and LAP are insufficient: LAP by itself cannot guarantee action-centric representations, while MCP alone misses visual dynamics. \emph{Our joint alignment bridges this gap, fusing the motion patterns from hidden states with the visual dynamics from latent actions.}
This dual constraint anchors the predictive embedding in a unified latent action space for hybrid data.

\subsection{Joint Perceivers with Decoupled Updates}
\label{sec:joint_perceiver}

To capture chunk-level dynamics more effectively, we use features from a pretrained visual encoder (e.g., DINOv3~\citep{simeoni2025dinov3} or V-JEPA2~\citep{bardes2024vjepa}) as inputs to our latent action module.
However, since these visual features originate from distinct visual backbones from the pretrained VLM, they may reside in a misaligned representation space.
Simply connecting them to predictive embeddings could lose important information or create latent actions that don't match the model's context.
To solve this, we introduce an additional \textbf{Latent State Perceiver (LSP)} to pair with the \textbf{Latent Action Perceiver (LAP)} (Fig.~\ref{fig:JALA_arch}).
While LAP processes boundary-frame features to generate latent actions, LSP connects the VLM's predictive context to the same latent action space to alleviate the predictive burden.

LAP and LSP share an identical 2-layer Perceiver architecture (with shared weights). Each layer contains a \textbf{cross-attention} block that attends from learnable latent queries to visual features (as key-value) and a \textbf{self-attention} block over the extracted latent tokens, followed by a 2-layer MLP that projects the latents into the VLM embedding space. To handle two-hand videos without sacrificing sharing, we use a shared Perceiver trunk but employ a two-head MLP by doubling its channel dimension and splitting it into left/right hand-specific heads, selected dynamically based on the active motion stream. For architectural consistency, both LAP and LSP operate on frame pairs: LAP takes the motion chunk boundary frames $(v_t, v_{t+\delta})$, while LSP uses a duplicated initial frame $(v_0, v_0)$, so that their differences arise from input semantics (dynamics vs.\ context) rather than architectural mismatch. 

A central challenge in aligning predictive embeddings with latent actions is that the Latent Action Perceiver (LAP) and Latent State Perceiver (LSP) process heterogeneous signals (visual boundary frames vs.\ predictive context), so directly coupling them with an alignment loss can be unstable or collapse if one side dominates. To stabilize training, we decouple optimization of the Perceiver \emph{backbone} and its learnable \emph{queries}, and adopt an asymmetric EMA update between LAP and LSP. Concretely, we optimize the backbone using gradients from LSP so that visual features are mapped into the predictive embedding space shaped by MCP and alignment, while optimizing the queries using gradients from LAP to anchor latent actions with explicit action cues. We then propagate backbone weights from LSP to LAP and query weights from LAP to LSP via:
\begin{align}
\theta^{\text{LAP}}_{b} &\leftarrow \alpha \, \theta^{\text{LAP}}_{b} + (1-\alpha) \, \theta^{\text{LSP}}_{b}, \\
\theta^{\text{LSP}}_{q} &\leftarrow \alpha \, \theta^{\text{LSP}}_{q} + (1-\alpha) \, \theta^{\text{LAP}}_{q},
\end{align}
where $\theta^{\text{LAP}}_{b}, \theta^{\text{LAP}}_{q}$ and $\theta^{\text{LSP}}_{b}, \theta^{\text{LSP}}_{q}$ denote the backbone/query parameters of LAP and LSP, respectively, and $\alpha \in [0,1)$ is the EMA coefficient. This asymmetric design keeps the LAP backbone consistent with the predictive context while allowing LSP queries to gradually inherit action-grounding capability, ensuring latent actions remain both predictable from context and anchored with action cues.

\subsection{Masked Chunk Prediction (MCP) Details}

\paragraph{Hybrid Masking Scheme.} During pretraining, we apply masked decoding to enforce predictive consistency.  
A naive masking strategy introduces a mismatch between training and inference, since all tokens in a chunk are replaced by \texttt{[MASK]} during training, but in inference, motion chunks are generated sequentially. To mitigate this gap, we employ a \textbf{hybrid masking scheme}:
\begin{itemize}[leftmargin=2em]
    \item For each sequence with $N$ chunks, we randomly select one chunk as the main prediction target.  
    \item Chunks before the target are kept intact (no masking).  
    \item Inside the target chunk, each token is masked with a random ratio uniformly sampled from $\{0.05, 0.15, \dots, 1.0\}$.  
    \item Tokens in chunks after the target are masked with a fixed 5\% probability to provide additional supervision without distorting context.  
\end{itemize}
This ensures that the main prediction chunk processes aligned context during both training and inference.  

\paragraph{Full Masking on Unlabeled videos.}
For in-the-wild \emph{video-only} samples that lack motion tokens, the entire motion chunk $A_i$ is replaced by \texttt{[MASK]} placeholders. In this case, the MCP term is inactive, and training proceeds solely via alignment to latent actions from LAP (i.e., only $\mathcal{L}_{\text{Align}}$ is applied). This keeps the interface unified while still learning predictive embeddings that are aligned to dynamics without requiring explicit motion labels.

\paragraph{Inference with MCP.}
For motion generation, we decode the current chunk \textbf{multiple times}, each time decoding $\sim$5\% of the tokens in the chunk. Finally, the outputs are ensembled to reduce approximation error.
This retains the efficiency advantage over causal decoding, while downstream transfer still uses a \textbf{single forward pass} to extract predictive embeddings.

\subsection{Two-Phase Training}

\paragraph{Pretraining on Hybrid Data.}
Once we establish predictive embeddings as a bridge to latent actions, we can use the same approach for videos without hand tracking labels.
For unlabeled videos, we skip MCP since no labels exist and only align predictive embeddings with visual dynamics from LAP.
This allows in-the-wild human manipulation videos to contribute useful learning signals despite lacking precise annotations.
As a result, we combine both labeled and unlabeled data using a hybrid training objective.
\begin{equation}
\mathcal{L} 
= \mathbf{1}_{\text{labeled}} \cdot \mathcal{L}_{\text{MCP}} 
+ \lambda \mathcal{L}_{\text{Align}},
\label{eq:hybrid_loss}
\end{equation}
where $\mathbf{1}_{\text{labeled}}$ is an indicator that only activates MCP when hand tracking labels are available. 
This means predictive embeddings learn from both sources: motion labels (when available) and dynamics (always), allowing JALA to scale pretraining across heterogeneous data. 

\paragraph{Post-training with Flow Matching.}
After pretraining on hybrid human data, JALA establishes a unified latent action space with aligned predictive embeddings. 
For robot manipulation tasks, we transfer this space to robot-specific actions with a \textbf{flow-matching head} that adapts predictive embeddings to the robot action space (Fig.~\ref{fig:JALA_arch}, right).
Given predictive embeddings $\{h_{i,k}\}$ from the pretrained VLA backbone, we feed them as a conditional input into the policy head, which is based on a Diffusion Transformer (DiT) with alternating self-attention and cross-attention modules. 
Self-attention processes the robot's proprioceptive state and noised actions, while cross-attention fuses this with predictive embeddings $\{h_{i,k}\}$, injecting the general dynamics knowledge learned from pre-training into action generation.
During post-training, we employ a flow-matching objective.
For a ground-truth action chunk $A_t$, we construct a noised action $A^\tau_t = \tau A_t + (1-\tau)\epsilon$ based on a timestep $\tau \in$ and standard Gaussian noise $\epsilon$.
The model $V_\theta$ learns to predict the denoising vector field $\epsilon - A_t$:
\begin{equation}
\mathcal{L}_{\text{FM}} = \mathbb{E}_{\tau, \epsilon, A_t} \left[ \left\| V_\theta(\{h_{i,k}\}, A^\tau_t, q_t) - (\epsilon - A_t) \right\|^2_2 \right],
\end{equation}
where $q_t$ is the robot's proprioceptive state.
During inference, by applying the learned model $V_\theta$, we iteratively denoise a random initial action chunk using forward Euler integration with a fixed step number. 
This ultimately yields precise robot action commands, enabling JALA to efficiently transfer dynamics-rich knowledge from human data to downstream precise robot control.

%% file: sections_arxiv/05_dataset.tex
\section{Hybrid Human Manipulation Dataset}
\label{sec:dataset}

To scale JALA pretraining, we construct \textbf{UniHand-Mix}, a hybrid human manipulation dataset that unifies \textbf{lab-collected annotated data} and \textbf{in-the-wild egocentric videos}.  
The design goal is to combine the strengths of both sources: lab data provides accurate hand-motion supervision and dense task annotations, while in-the-wild data contributes broader behavioral diversity and richer contextual priors.

Overall, UniHand-Mix consists of:
\begin{itemize}[leftmargin=1.5em]
    \item a \textbf{lab-annotated subset} with paired instructions, videos, and MANO-based hand motion sequences;
    \item an \textbf{in-the-wild subset} curated from Ego4D~\citep{grauman2022ego4d}, with automatically validated hand-centric clips and generated instructions, where a portion also includes pseudo hand-pose annotations.
\end{itemize}

In total, UniHand-Mix contains over \textbf{7.5M} instruction--video samples, including \textbf{5M+} lab-annotated instruction--video--motion samples and \textbf{2.5M} in-the-wild instruction--video pairs (about \textbf{10\%} with hand-pose annotations).

\subsection{Lab-Annotated Subset}
\label{subsec:lab_subset}

Following the UniHand pipeline~\cite{beingbeyond2025beingh0}, we curate a high-quality lab-collected subset with precise 3D hand supervision and dense task descriptions. Each sample contains an instruction, a video clip, and a MANO-based hand motion sequence. We further support three instruction-tuning task types: \textit{motion generation}, \textit{motion description}, and \textit{motion continuation}.

\paragraph{Data curation pipeline.}
We build the lab subset through a three-step pipeline: hand-pose standardization, hierarchical task labeling, and instructional sample generation. Together, these steps convert heterogeneous human manipulation datasets into a unified vision--language--motion training corpus:
\begin{itemize}[leftmargin=2em]
    \item[] \textbf{Hand pose standardization.}
    We unify all hand annotations into the MANO parameter format~\citep{romero2017mano}.  
    For datasets with mocap or SLAM labels, we directly convert annotations to MANO.  
    For datasets with only 3D joints, we fit MANO parameters via optimization.  
    For RGB-only videos, we estimate per-frame hand pose using HaWoR~\citep{zhang2025hawor}, followed by temporal smoothing and left--right correction.

    \item[] \textbf{Hierarchical task labeling.}
    We segment videos into 10-second clips and annotate them hierarchically.  
    At the clip level, we provide imperative instructions and concise summaries.  
    At the second level, we annotate contact states, object properties, and hand--object interactions, covering both bimanual and single-hand actions.

    \item[] \textbf{Instructional data generation.}
    Based on the annotations above, we construct multimodal training samples for motion generation, motion description, and motion continuation.  
    We start from task-specific base templates and use Gemini to diversify the language, improving linguistic coverage while preserving alignment between vision, language, and motion.
\end{itemize}

Using this pipeline, we process approximately \textbf{1,000 hours} of lab videos, yielding \textbf{5M+} instruction--video--motion samples.

\subsection{In-the-Wild Subset}
\label{subsec:wild_subset}

To complement the lab subset with more diverse and natural human behaviors, we curate an in-the-wild subset from Ego4D~\citep{grauman2022ego4d}. Compared with controlled lab environments, egocentric videos are substantially noisier: hand visibility is often intermittent due to occlusion, and many clips contain idle or non-manipulative hand motion. To distill usable manipulation episodes, we design a two-stage filtering and annotation pipeline.

\paragraph{Data curation pipeline.}
We adopt a staged filtering-and-annotation pipeline to extract reliable manipulation episodes from noisy egocentric videos. The pipeline progressively improves sample quality through visual filtering, hand-centric activity validation, and optional pseudo hand-pose annotation:
\begin{itemize}[leftmargin=2em]
    \item[] \textbf{Visual filtering.}
    We apply WiLoR~\citep{potamias2025wilor} for frame-level hand-region detection and remove clips without visible human hands.

    \item[] \textbf{Hand-centric activity validation and instruction generation.}
    For the remaining clips, we use \texttt{Gemini-2.5-Flash} to identify hand-centric manipulation activities, discarding samples with absent, idle, or distractor hands.  
    For valid clips, we automatically generate paired instructions.

    \item[] \textbf{Pseudo hand-pose annotation (optional).}
    We further estimate hand pose using HaWoR~\citep{zhang2025hawor} and retain high-confidence clips (confidence threshold 0.65).  
    These clips provide approximate hand supervision and serve as an intermediate bridge between fully annotated lab data and video-only in-the-wild data.
\end{itemize}

This process yields approximately \textbf{2.5M} instruction--video pairs, of which about \textbf{10\%} include pseudo hand-pose annotations. The resulting subset significantly expands task diversity and contextual coverage, improving pretraining scalability.

\begin{figure}[t]
    \centering
    \includegraphics[width=0.81\linewidth]{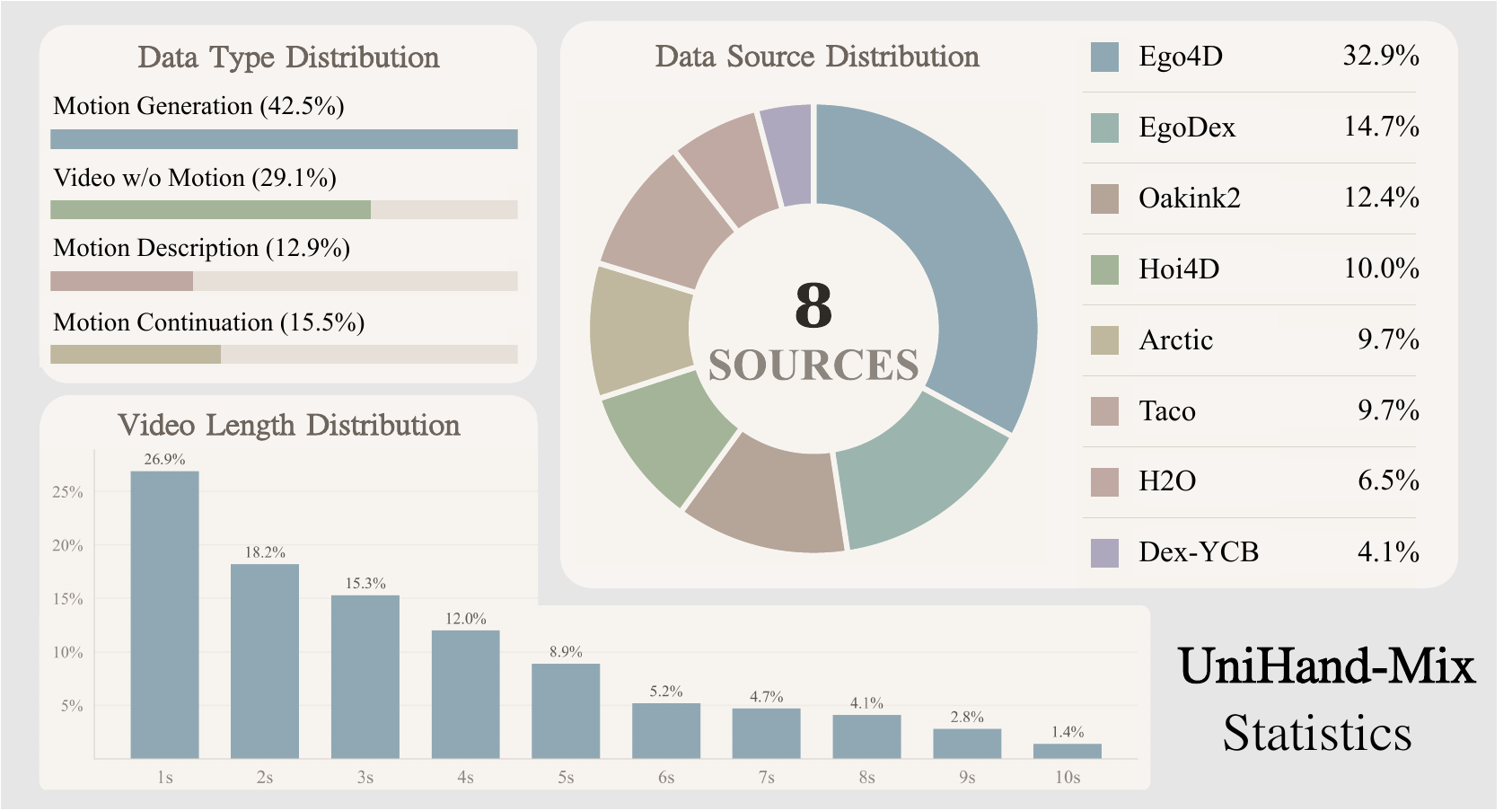}
    % \vspace{-6pt}
    \caption{\textbf{Dataset statistics of UniHand-Mix.}
    Top-left: distribution of data types (motion generation, video-only, motion description, and motion continuation).
    Bottom-left: distribution of clip lengths (1--10 seconds).
    Center: data source distribution across 8 data sources with a donut percentage chart on the right.}
    \label{fig:data_stats}
    % \vspace{-4pt}
\end{figure}

\subsection{Statistics}
\label{subsec:dataset_stats}

Figure~\ref{fig:data_stats} summarizes UniHand-Mix along three axes: task type, clip length, and data source. Overall, the dataset is designed to balance \emph{high-quality motion supervision} from lab-collected data with \emph{large-scale behavioral diversity} from in-the-wild videos, while maintaining broad temporal coverage for different training objectives.

\paragraph{Task-type composition.}
The task distribution is intentionally heterogeneous rather than uniform.
\textit{Motion generation} forms the largest portion and serves as the main supervision signal for learning visual-to-motion mapping.
At the same time, a substantial fraction of \textit{video-only} samples is retained to expand visual and contextual diversity beyond hand-annotated data.
The remaining \textit{motion description} and \textit{motion continuation} samples provide complementary supervision for language alignment and temporal modeling.

\paragraph{Temporal coverage.}
The clip-length distribution is skewed toward shorter clips, with progressively fewer long clips.
This reflects the structure of human manipulation data: short segments capture dense local interactions and fine-grained motion primitives, while longer clips preserve multi-step behaviors and temporal dependencies.
By mixing both regimes, UniHand-Mix provides supervision for both short-horizon motion learning and longer-horizon action continuation.

\paragraph{Source diversity.}
UniHand-Mix integrates eight sources spanning both lab-collected datasets and Ego4D in-the-wild videos.
The in-the-wild portion contributes a large share of the dataset, ensuring broad coverage of environments, objects, and interaction contexts.
Meanwhile, the lab-collected sources remain a substantial component, anchoring the dataset with more reliable hand-motion supervision and cleaner manipulation episodes.
The lab subset itself is distributed across multiple benchmarks rather than concentrated in a single source, which improves diversity in capture setups, tasks, and object categories.

Overall, the statistics suggest three desirable properties of UniHand-Mix for pretraining: 
(1) strong motion supervision from lab-collected data, 
(2) broad contextual diversity from in-the-wild and video-only samples, and 
(3) multi-scale temporal coverage from short and long clips.
These properties jointly support scalable JALA pretraining across motion prediction, language grounding, and temporal reasoning.

%% file: sections_arxiv/06_experiments.tex
\section{Experiments}
\label{sec:experiments}
Our experimental study investigates three core questions:
\textbf{(1) In-the-wild Learning:}
Can JALA learn from in-the-wild human videos and demonstrate scaling?
\textbf{(2) Downstream Transfer.}
Does the pretrained predictive embedding improve robot task performance?
\textbf{(3) Ablation Validation.}
Are the proposed alignment designs effective?

\subsection{Implementation Details}

\noindent\textbf{Pre-training.}
JALA follows the Being-H0 design and is built on InternVL3-2B~\citep{chen2024internvl} as the vision-language backbone of 28 attention layers, with DINOv3~\citep{simeoni2025dinov3} or V-JEPA2~\citep{bardes2024vjepa} as the visual encoder feeding into the latent perceiver.
To incorporate the motion modality, we tokenize \textit{15-frame motion chunks}. Each chunk is decomposed into \textit{wrist} and \textit{finger} motions, which are separately quantized into 64 tokens each, yielding 128 tokens per chunk.  The codebook size of each part is 4096, learned via \textit{GRVQ}~\citep{yang2023hifi}, a vector quantization algorithm, to capture general motion patterns.
The motion tokens are then wrapped with two \textbf{special tokens} \texttt{<mot>} and \texttt{</mot>}, forming the following unified format:
\[
\texttt{<mot>}\;\{\text{wrist\_motion\_tokens}\}\;\{\text{finger\_motion\_tokens}\}\;\texttt{</mot>}
\]
These special tokens serve as explicit delimiters, helping the VLM distinguish motion chunks from other modalities such as vision or instruction tokens.
For sequences containing \textbf{both hands}, we interleave the left-hand and right-hand chunks along the temporal axis to preserve synchrony while maintaining modality distinction.
In-the-wild data is temporally slowed by a factor of $0.5$, to account for the differing action speeds with lab-collected data.
Predictive embeddings are taken from the 19th attention layer (out of 28) for joint alignment with inverse-dynamics signals. The training objective combines masked chunk prediction and latent-action alignment with $\lambda=0.5$. For in-the-wild videos, only the alignment loss is applied. We use AdamW with base learning rate $3 \times 10^{-5}$, weight decay $0.05$, and $\beta=(0.9,0.95)$. The learning rate is warmed up for the first 5\% steps and then decayed with a cosine schedule. Gradient clipping with max norm 1.0 is applied throughout. The perceiver modules are updated with an EMA coefficient of $\alpha=0.999$. Training uses an effective batch size of 128 sequences, obtained from a per-GPU batch size of 16 with gradient accumulation across 8 GPUs, where each sequence contains a 15-frame motion chunk plus paired instructions and boundary frames. Pretraining is performed for a single epoch on the full 7.5M UniHand-Mix dataset, requiring 68 hours on 8 NVIDIA A800 (80GB) GPUs.

\noindent\textbf{Post-training.}
After pre-training, we fine-tune JALA on LIBERO, RoboCasa and real-world tasks using a flow-matching head. We implement a Diffusion Transformer (DiT) policy head, which is composed of alternating self-attention and cross-attention layers.  We employ a DiT with 16 layers of 32-head attention blocks, and the hidden state dimension is 2048.
Training uses batch size 128 with learning rate $1 \times 10^{-4}$, 5\% warm-up, and cosine decay. Using the same hardware configuration as in pretraining, we run 30k steps ($\approx$8 hours) on LIBERO, 60k steps ($\approx$16 hours) on RoboCasa, and 30k steps ($\approx$8 hours) on real-world tasks.
Only language model parameters are unfrozen, and the vision encoder remains frozen.
We train a GR00T-N1.5 baseline for comparison with identical settings. During inference, we use $N=4$ denoising steps.

\subsection{Hand Motion Generation}

We evaluate JALA on hand motion generation, which predicts a motion chunk given an initial visual input and instruction.
This task serves as a direct indicator of how well a model internalizes human manipulation priors from pretraining data.
More importantly, it tests whether JALA's joint-aligned latent actions form a unified latent action space that supports both precisely annotated data and unconstrained, unlabeled in-the-wild videos.
\input{tables/motion_gen}

\paragraph{Model Variants and Baselines.}
We evaluate two variants of JALA: \textbf{JALA-dino} and \textbf{JALA-vjepa}, which differ only in the visual backbone used for LAP/LSP (DINOv3 vs.\ V-JEPA2).
For comparison, we include four baselines.
(1) \textbf{Being-H0}~\citep{beingbeyond2025beingh0}, our reproduction using the same \texttt{InternVL3-2B} backbone.
(2) \textbf{Being-H0+dino}, which adds DINO features to Being-H0 to match JALA's LSP input.
(3) \textbf{JALA w/o align}, which removes LAP entirely to isolate the contribution of joint alignment.
(4) \textbf{JALA w/o latent}, which first trains an action-token predictor mapping boundary frames to motion chunk tokens, and uses its predicted action tokens as explicit pseudo-labels on video-only data as an alternative to latent actions for supporting hybrid dataset.
JALA-dino, JALA-vjepa, and JALA w/o latent are all trained on the full UniHand-Mix, whereas the remaining baselines are trained only on subsets where action labels are available.

\noindent\textbf{Evaluation metrics.}  We report results on two evaluation splits of UniHand-Mix:
\textbf{(1) Lab split}, a held-out subset from the lab-annotated data that measures fidelity under precise supervision;
\textbf{(2) Wild split}, curated from Ego4D with HaWoR annotations that measures generalization to unseen and in-the-wild manipulation behaviors.
We report four metrics:
\textbf{(1) MPJPE}, the mean Euclidean distance between predicted and ground-truth 3D joints for spatial accuracy measurement;
\textbf{(2) PA-MPJPE}, the MPJPE after rigid alignment for relative pose fidelity measurement;
\textbf{(3) MWTE}, the average offset of wrist trajectories for global trajectory fidelity measurement;
\textbf{(4) MDE}, the error of final displacement direction from the initial wrist position for motion trend consistency measurement.
These metrics capture both local pose accuracy and global trajectory realism.

\paragraph{Dataset setups.}
We follow the same protocol as in pretraining for motion generation evaluation.
The \textbf{Lab split} is a held-out subset of lab-annotated sequences that measures fidelity under precise supervision.
The \textbf{Wild split} is curated from Ego4D videos with HaWoR annotations to evaluate generalization to unconstrained, in-the-wild manipulation behaviors.
These two splits reflect different regimes: controlled indoor environments versus diverse real-world activities.

\noindent\textbf{Main Results.}
Tab.~\ref{table:hand_motion_generation} shows that both JALA variants outperform baselines on almost all metrics.
Gains are modest on the Lab split, but substantial promotion is shown on the Wild split. While all models degrade on the Wild split compared to the Lab split due to higher variability and complexity, the performance drop is much smaller on JALA variants. This demonstrates that joint-aligned latent actions do capture manipulation priors that transfer beyond curated lab settings, enabling robust scaling to realistic human interactions.

\noindent\textbf{Design Insights.}
The ablations reveal several insights into why JALA works.
(1) Comparing Being-H0+dino with Being-H0 shows that adding stronger visual features provides only mild improvements. (2) The similar performance of JALA w/o align indicates that the prediction paradigm (next-token vs.\ masked-chunk) is not the primary factor. (3) JALA w/o latent performs noticeably worse than JALA w/o align, particularly on the Wild split, even though it is trained with more data. This gap shows that simply increasing the unlabeled video amount can be detrimental without appropriate structural supervision. (4) Joint-aligned latent actions instead offer stable and transferable information, allowing JALA to benefit from large-scale in-the-wild video rather than being hindered by it. The similar performance of JALA-dino and JALA-vjepa further suggests that JALA is robust to the choice of self-supervised visual backbone. Overall, the comparisons suggest that adjustments to prediction paradigms, visual features, or data volume play a limited role, whereas joint-aligned latent actions constitute the primary driver of JALA's generalization gains.

\begin{figure}[t]
    \centering
    \includegraphics[width=0.45\linewidth]{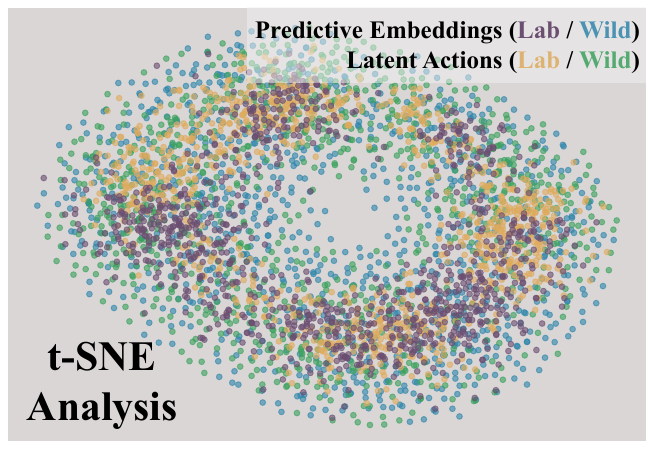}
    \vspace{-6pt}
    \caption{\textbf{t-SNE of predictive embeddings $h$ and latent actions $z$ across Lab and Wild.}
    The two spaces cluster in closely aligned regions, and Wild samples largely expand the Lab manifold.}
    \label{fig:tsne}
    \vspace{-4pt}
\end{figure}

\begin{figure}[t]
    \centering
    \includegraphics[width=0.8\linewidth]{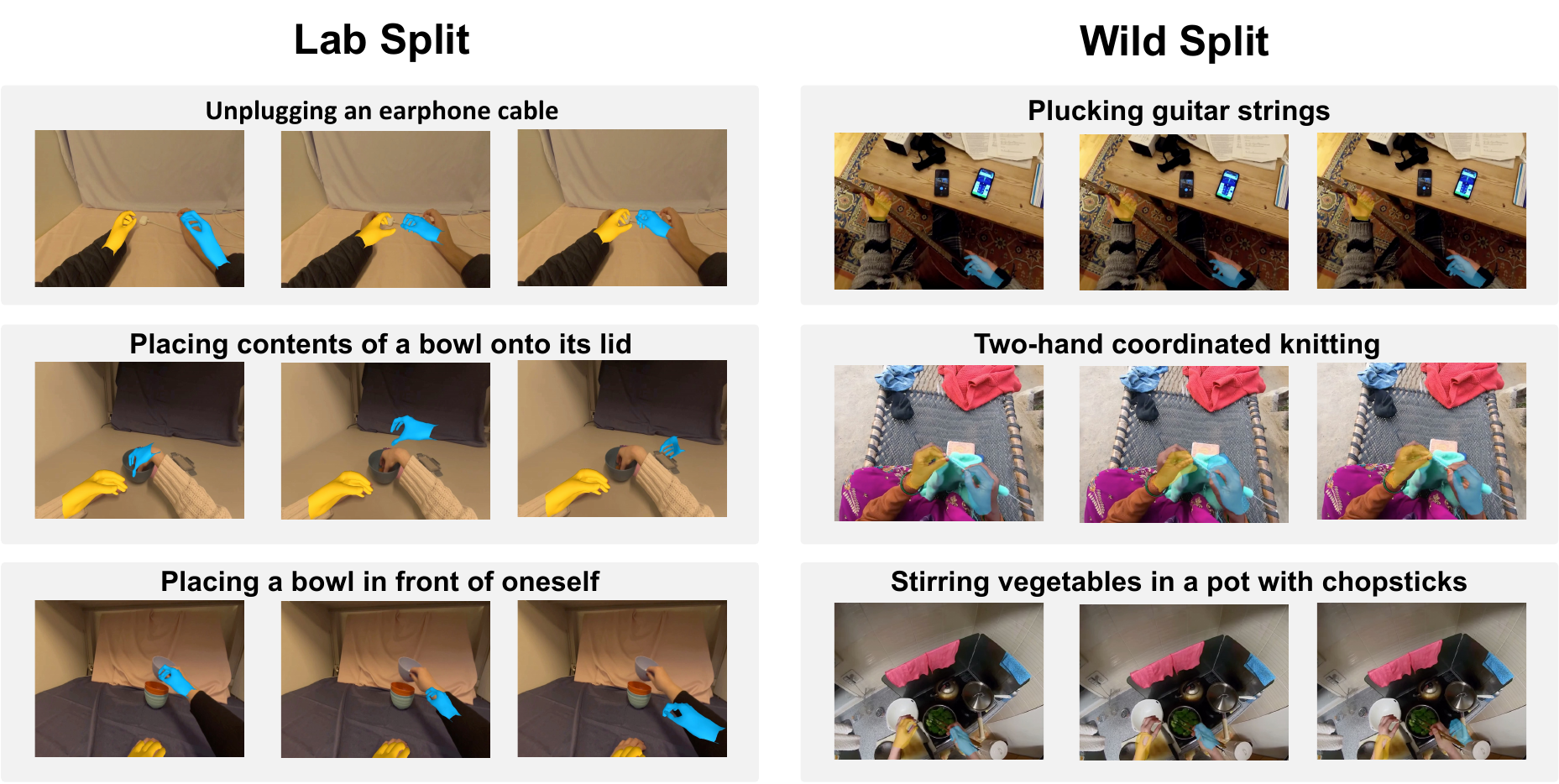}
    \caption{\textbf{Qualitative hand-motion generation on lab (left column) and wild (right column) scenes.}
    Colored overlays denote generated hand poses.}
    \label{fig:qual_wild_lab}
\end{figure}

\noindent\textbf{Visualization.}
To further verify that joint alignment produces a coherent latent space, we visualize the alignment between predictive embeddings $h$ and latent actions $z$ across Lab and Wild data. As shown in Fig.~\ref{fig:tsne}, the two spaces cluster in closely aligned regions, and Wild samples largely expand the Lab manifold, indicating integrated coverage rather than a disjoint domain. This confirms that JALA successfully bridges the gap between curated lab data and unconstrained in-the-wild videos within a single latent action space.
Figure~\ref{fig:qual_wild_lab} further presents qualitative hand-motion generations covering both in-the-wild and lab-collected scenarios. On the wild side, the model successfully handles diverse, unconstrained interactions such as plucking guitar strings, two-hand coordinated knitting, and stirring vegetables in a pot with chopsticks, demonstrating robust generalization to complex scenes and bimanual coordination. On the lab side, the model produces precise and temporally consistent motions for unplugging an earphone cable, placing contents of a bowl onto its lid, and placing a bowl in front of oneself, reflecting accurate fine-grained control in structured settings. These qualitative results corroborate the quantitative trends in Tab.~\ref{table:hand_motion_generation}.

\subsection{Simulation Results}

To evaluate downstream transfer, we fine-tune JALA on three simulation benchmarks: LIBERO, RoboCasa, and GR1 tabletop tasks. Before presenting per-benchmark results, we first introduce the additional baselines and ablations that are shared across these experiments.

\paragraph{Additional baselines and ablations.}
Beyond the baselines already used in hand motion generation, we introduce five new comparisons for the simulation experiments.
(1) \textbf{LAPA}~\citep{ye2024lapa} (official checkpoint), a reconstruction-based latent action method.
(2) \textbf{LAPA$^\dagger$}, LAPA retrained on our data with the JALA backbone to isolate the effect of the training objective from backbone and data differences.
(3) \textbf{JALA$^\star$}, a variant of JALA that treats the annotated wild clips as unlabeled, testing whether a small ratio of wild motion annotations is essential.
(4) \textbf{JALA-act}, JALA trained only on the action-available subset, to disentangle the contribution of the method from the additional data.
(5) \textbf{JALA w/o dec.}, JALA-act without the decoupled EMA update, to verify the necessity of the stabilization mechanism.
These comparisons allow us to systematically disentangle the effects of training objective (reconstruction vs.\ joint alignment), data composition (action-available vs.\ full UniHand-Mix), and architectural design (with vs.\ without decoupled EMA).

\subsubsection{LIBERO Experiments}

\noindent\textbf{Setup.}
We independently fine-tune our models on the four task suites (Spatial, Object, Goal, and Long) from LIBERO.
We follow the original two-view setting and further evaluate a more challenging \emph{single third-person view} setting that removes the auxiliary egocentric camera and therefore requires stronger abstraction from visual observations.
We compare JALA with state-of-the-art VLA models under both camera setups.
Specifically, we include LAPA and UniVLA in our baselines, which utilize reconstruction-based latent action for video-only pretraining, for a comparison of joint-aligned latent action and reconstruction-based latent action.
For the single-view experiments, we additionally include JALA w/o align and JALA w/o latent to validate our design choices.
All models are fine-tuned with fewer than 50 demonstrations per task.
\input{tables/libero_2}

\noindent\textbf{Two-View Results.}
Under the standard two-view setting, recent VLA models already achieve strong performance, including the reconstruction-based method UniVLA.
As shown in Table~\ref{table:libero_two_view}, JALA performs on par with or better than these stronger baselines despite being lighter.
JALA-dino attains the highest overall average success rate of 96.9\% without any pretraining on robotic data, indicating that joint-aligned latent actions provide valid action abstraction generalizable to downstream robotics tasks.
Both reconstruction-based baselines, LAPA (official checkpoint, 79.5\%) and LAPA$^\dagger$ (retrained on our data with the same backbone, 83.5\%), fall clearly behind JALA. Although LAPA$^\dagger$ improves over the original LAPA by benefiting from our data and backbone, it still trails JALA by over 13 points on average, suggesting that the performance gap stems from the training objective rather than data or architecture differences.
On the same action-available subset, JALA-act (94.3\%) consistently improves over Being-H0 (90.2\%), with the largest gain on the Long suite (91.8\% vs.\ 77.4\%), supporting the benefit of joint alignment for learning transferable human priors even without in-the-wild data.
JALA$^\star$, which treats annotated wild clips as unlabeled, remains close to full JALA (95.7\% vs.\ 96.9\%), indicating that a small ratio of motion annotations in wild data helps but is not essential for strong performance.
The sharp collapse of JALA w/o dec.\ (56.6\%) highlights the critical role of the decoupled EMA update in stabilizing alignment. Without it, the LAP and LSP modules fail to converge to a coherent latent space.

\input{tables/libero}
\noindent\textbf{Single-View Results.}
As shown in Table~\ref{table:libero}, both of our JALA variants still demonstrate top-tier performance in this harder setting.
JALA-dino achieves a new SoTA average success rate of 92.3\% among models with comparable backbone scale (<3B parameters), outperforming strong baselines such as GR00T N1.5, even though JALA is pretrained only on human data.
The gains are especially large on the Long suite, where JALA-dino achieves 87.2\%, demonstrating superior generalization capabilities on long-horizon tasks. The ablations highlight the importance of our design: removing joint alignment (JALA w/o align) yields a clear drop in performance, and simply adding up pretraining video data (JALA w/o latent) causes a much larger degradation, confirming that joint-aligned latent actions are critical to the JALA's outperformance. Importantly, reconstruction-based models such as UniVLA require significantly more computation, GPU memory, and training.  Despite this heavy resource usage, our JALA, without any reconstruction and using only human videos for pretraining, still surpasses the human-only UniVLA variant and approaches the performance of UniVLA-full, which additionally benefits from large-scale robot datasets. This demonstrates that joint-aligned latent actions provide a more efficient route for extracting transferable manipulation priors from in-the-wild human videos.

\subsubsection{RoboCasa Experiments}

\noindent\textbf{Setup.}
RoboCasa provides a kitchen-environment benchmark with both synthetic and human-collected demonstrations. To assess the model's performance on different data sources, we conduct two experiments on the atomic tasks: the first one trained with 50 \emph{human demonstrations} per task, and another with 50 \emph{synthetic demonstrations} per task. We deliberately use only 50 demonstrations per task to match the human split (which provides only 50 per task) and to emphasize a few-shot downstream regime that is more reflective of prior generalization than large-demo finetuning.

\noindent\textbf{Results.}
As shown in Table~\ref{table:robocasa}, JALA consistently outperforms reconstruction-based baselines and ablations on both RoboCasa splits.
The advantage is particularly pronounced on synthetic data, where the domain gap between pretraining videos and downstream demonstrations is larger.
JALA-act again improves over Being-H0 on the same action-available subset, while JALA w/o dec.\ degrades sharply, confirming the necessity of decoupled EMA updates.
The consistent gap between LAPA$^\dagger$ and JALA, under identical backbone and data, further supports the advantage of joint alignment over reconstruction-based objectives in few-shot transfer scenarios.

\subsubsection{GR1 Tabletop Experiments}

\noindent\textbf{Setup.}
Unlike LIBERO and RoboCasa, which use parallel-jaw grippers, GR1 tabletop tasks employ a dexterous hand end-effector that is closer to the human hand morphology seen in our pretraining videos. This makes GR1 a particularly relevant testbed for evaluating whether human-derived manipulation priors transfer to a morphologically similar but distinct robotic embodiment. We use a single third-person camera view and a few-shot setting with 50 demonstrations per task.

\noindent\textbf{Results.}
As shown in Table~\ref{table:robocasa}, JALA achieves the highest success rate of 26.33\% on GR1, substantially outperforming GR00T N1.5 (20.41\%), LAPA (11.42\%), and Being-H0 (12.91\%).
The gain is especially notable compared with Being-H0. On the same action-available subset, JALA-act (20.25\%) nearly doubles Being-H0's score (12.91\%), demonstrating that joint alignment is particularly beneficial when the downstream embodiment diverges from the pretraining data.
LAPA$^\dagger$ (13.50\%) also trails JALA by a large margin despite using the same backbone and data, reinforcing that the training objective, not the data or architecture, drives the improvement.
% Error bars are omitted for RoboCasa and GR1 due to evaluation cost (each reported score aggregates $24\times 50=1200$ runs).

\input{tables/robocasa}

\begin{figure}
    \centering
    \includegraphics[width=\linewidth]{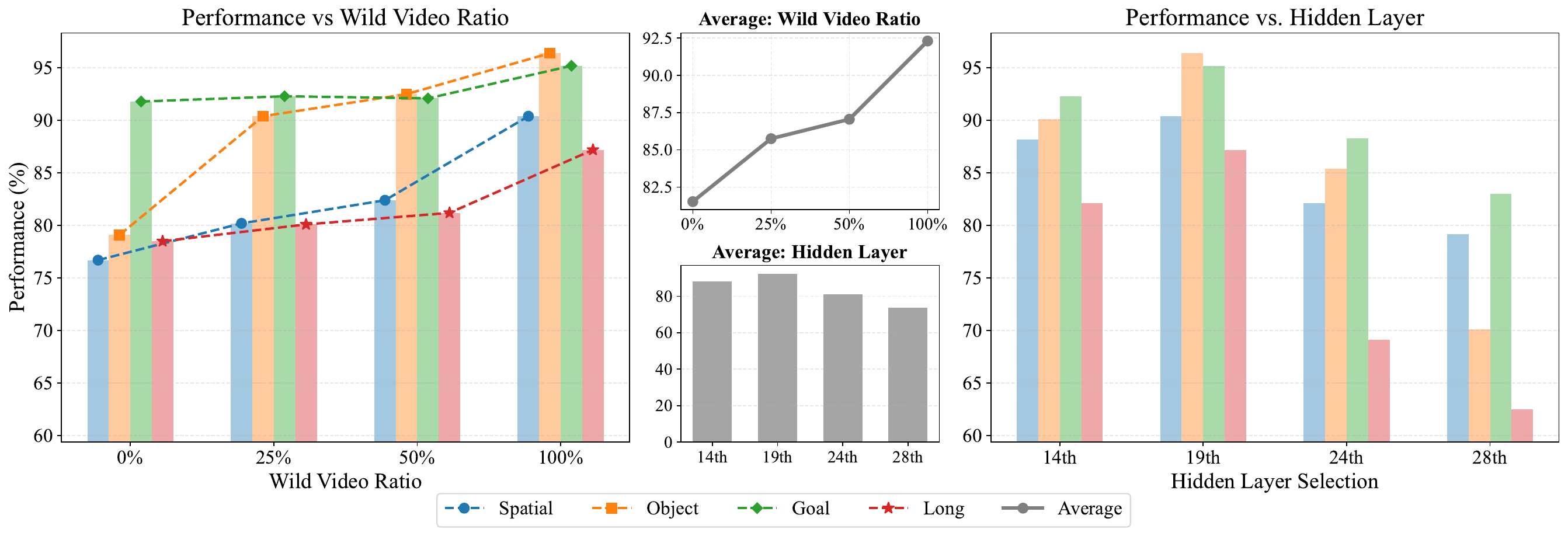}
    \caption{Ablation studies on JALA-dino evaluated on LIBERO.
    \textbf{Left:} performance across various the proportion of in-the-wild data used during pretraining (0\%, 25\%, 50\%, 100\%).
    \textbf{Right:} performance when feeding hidden states from different backbone layers (14, 19, 24, 28) into the flow-matching head during adaptation. }
    \label{fig:scaling}
\end{figure}
\subsubsection{Ablation Studies}
We conduct two ablations on the LIBERO benchmark with JALA-dino to validate (i) its scalability to unconstrained human videos, and (ii) the role of predictive embeddings compared with other hidden states.

\noindent\textbf{Scalability to in-the-wild data.}
To assess how large uncurated videos contribute to JALA, we vary the proportion of in-the-wild data during pretraining while keeping lab-annotated data fixed (0\%, 25\%, 50\%, and 100\%). Results on LIBERO (Fig.~\ref{fig:scaling}, left) show a consistent improvement in downstream success rates as more in-the-wild data are added, confirming the strong scaling potential of our framework.

\noindent\textbf{Flow-matching input ablation.}
We further test the effectiveness of different hidden states from the backbone when fed into the flow-matching head during downstream adaptation. Note that we only vary the layer providing inputs to flow-matching on the JALA-dino pre-trained with the 19\textsuperscript{th} layer used for alignment.
We compare hidden states from layers 14, 19, 24, and 28. As shown in Fig.~\ref{fig:scaling} (right), using the 19\textsuperscript{th} layer yields the best transfer performance, the 14\textsuperscript{th} layer is slightly worse, while later layers degrade sharply.
This might suggest that alignment concentrates generalizable cues in the selected layer, whereas deeper layers overfit to dataset-specific details, reducing their utility for downstream robot control.

\subsubsection{Reconstruction vs.\ Joint Alignment}
\label{sec:additional_results}

Across all three simulation benchmarks, we observe a consistent empirical gap between reconstruction-based baselines and joint alignment. Specifically, JALA outperforms LAPA and LAPA$^\dagger$ on two-view LIBERO (Table~\ref{table:libero_two_view}) and shows stronger gains on RoboCasa and GR1 (Table~\ref{table:robocasa}), including under embodiment shift and few-shot settings.
Reconstruction-based objectives such as LAPA provide dense pixel supervision, which can place more weight on appearance, background, and camera artifacts in in-the-wild videos. This is not always aligned with action-relevant dynamics, especially when hands are small or partially occluded. Joint alignment instead constrains predictive embeddings using boundary-frame dynamics, yielding a more behavior-centric signal that transfers better when data are scarce or domain shifts are large.
The controlled comparison with LAPA$^\dagger$, which shares the same backbone and data as JALA, isolates the effect of the training objective. The remaining performance gap therefore reflects differences between reconstruction and alignment, rather than backbone or data scale.
Beyond accuracy, the two paradigms also differ in training efficiency. With identical backbone and data, LAPA$^\dagger$ requires two-stage pretraining (29h + 57h) versus 68h for JALA on 8$\times$A800-80G, so JALA uses less than 80\% of the wall-clock time while achieving better performance. Pixel reconstruction provides dense targets but tends to amplify background shift or noise in in-the-wild videos, where much compute is spent on action-agnostic signals. In contrast, JALA offers a more behavior-centric supervision with action-relevant gradients from joint alignment, making it both more effective and more efficient.

\subsection{Real-World Robot Experiments}
\begin{figure}[t]
    \centering
    \includegraphics[width=0.85\linewidth]{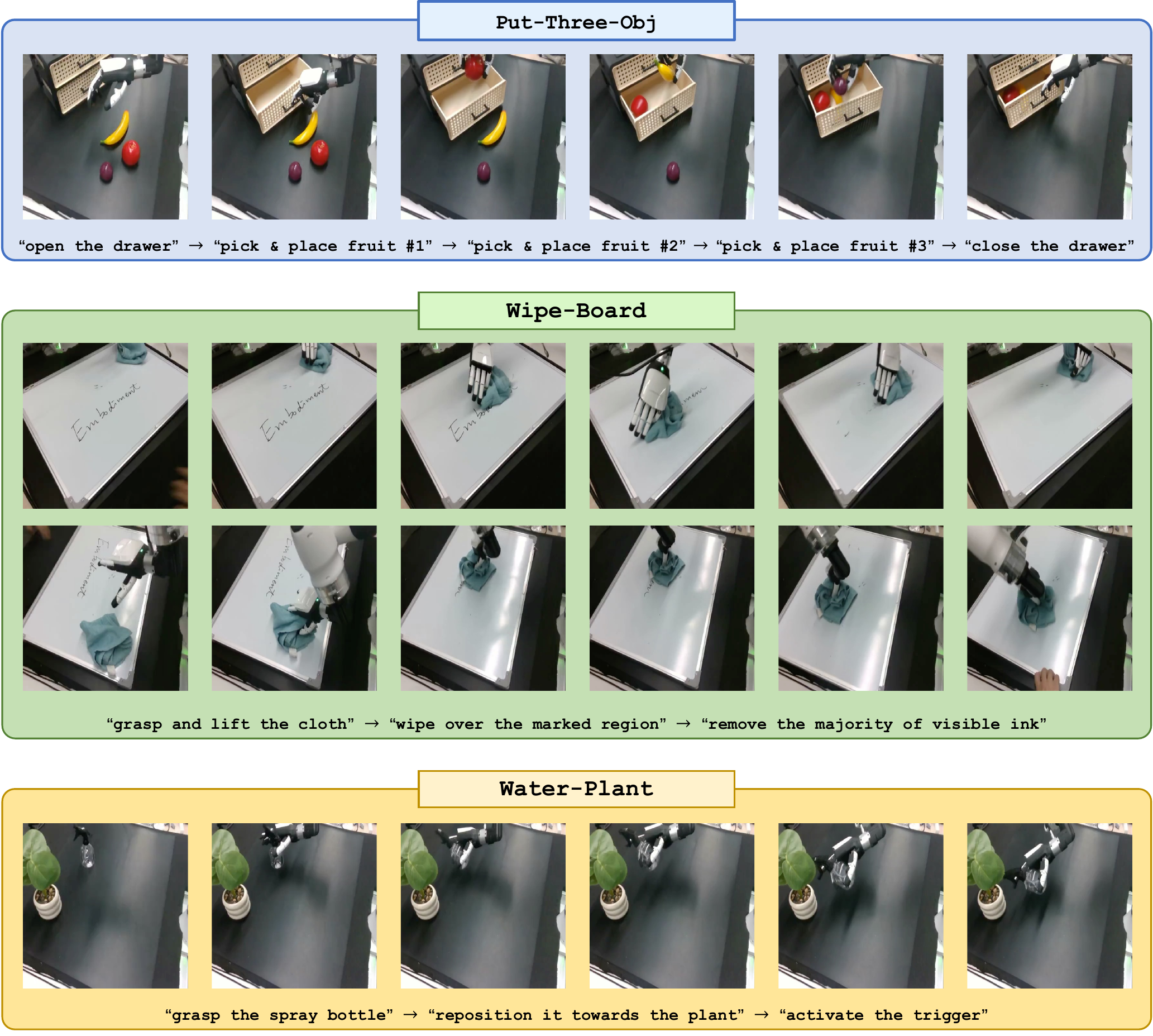}
    \caption{
    \textbf{Teleoperated demonstration examples} for the three real-world manipulation tasks.
    \textbf{Top}: \texttt{\textbf{Put-Three-Obj}} (open drawer $\to$ pick\&place three fruits $\to$ close drawer).
    \textbf{Middle}: \texttt{\textbf{Wipe-Board}} (grasp cloth $\to$ wipe over marked region $\to$ remove visible ink).
    \textbf{Bottom}: \texttt{\textbf{Water-Plant}} (grasp spray bottle $\to$ reposition toward plant $\to$ activate trigger).
    }
    \label{fig:teleop_examples}
\end{figure}
\begin{figure}
    \centering
    \includegraphics[width=0.8\linewidth]{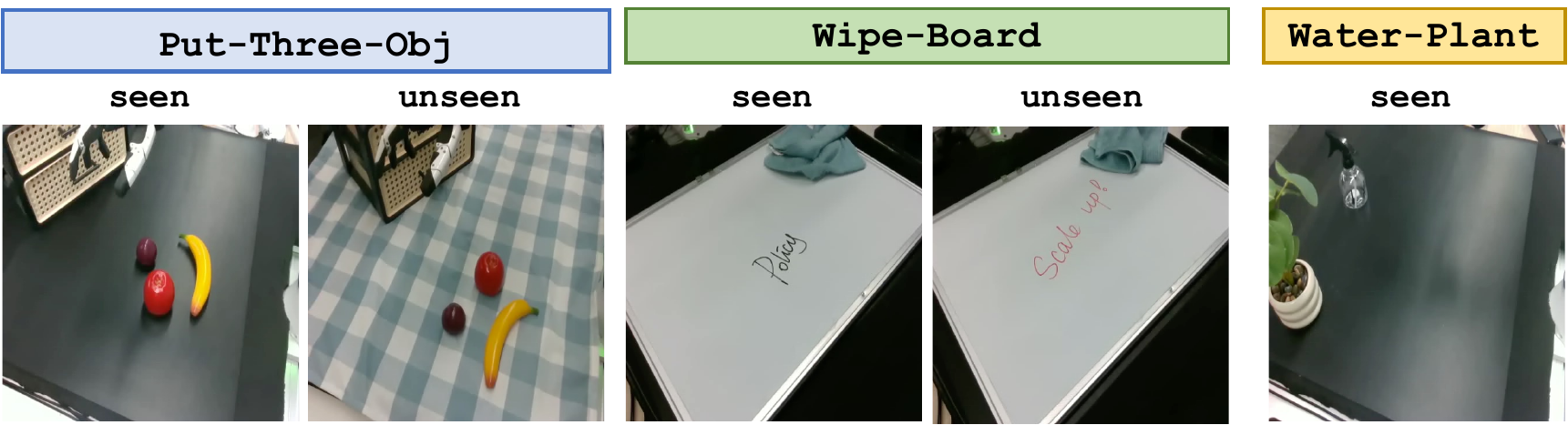}
    \caption{Real-world robot task settings for three multi-step tasks.
Put-Three-Obj and Wipe-Board include unseen variants to evaluate robustness to visual shifts.}
    \label{fig:setup}
\end{figure}

\begin{figure}[h]
    \centering
    \includegraphics[width=0.7\linewidth]{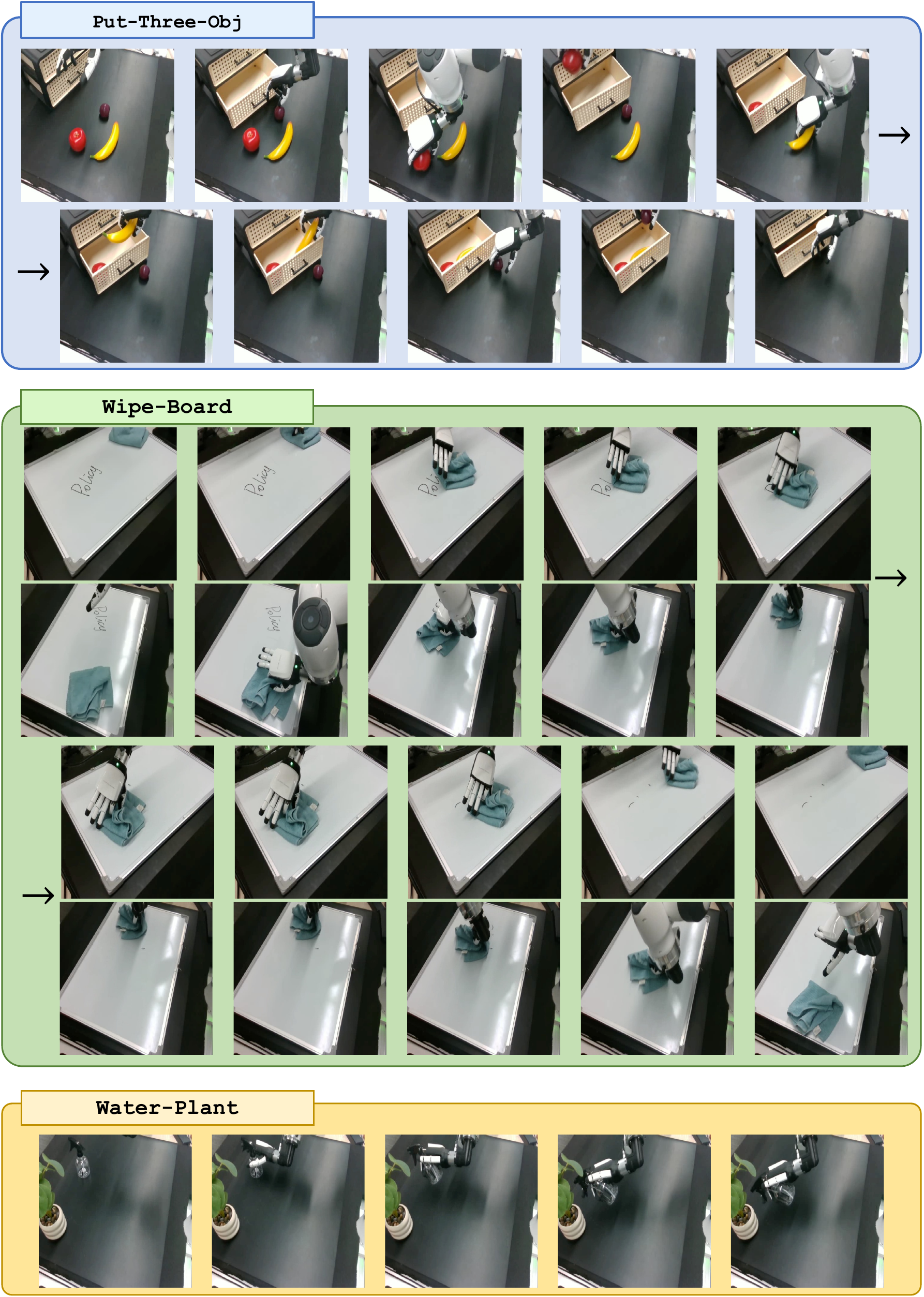}
    \caption{
    \textbf{Successful real-robot rollouts (seen setting).}
    Top: \texttt{Put-Three-Obj}; Middle: \texttt{Wipe-Board}; Bottom: \texttt{Water-Plant}.
   }
    \label{fig:success_seen}
\end{figure}

\begin{figure}[h]
    \centering
    \includegraphics[width=0.7\linewidth]{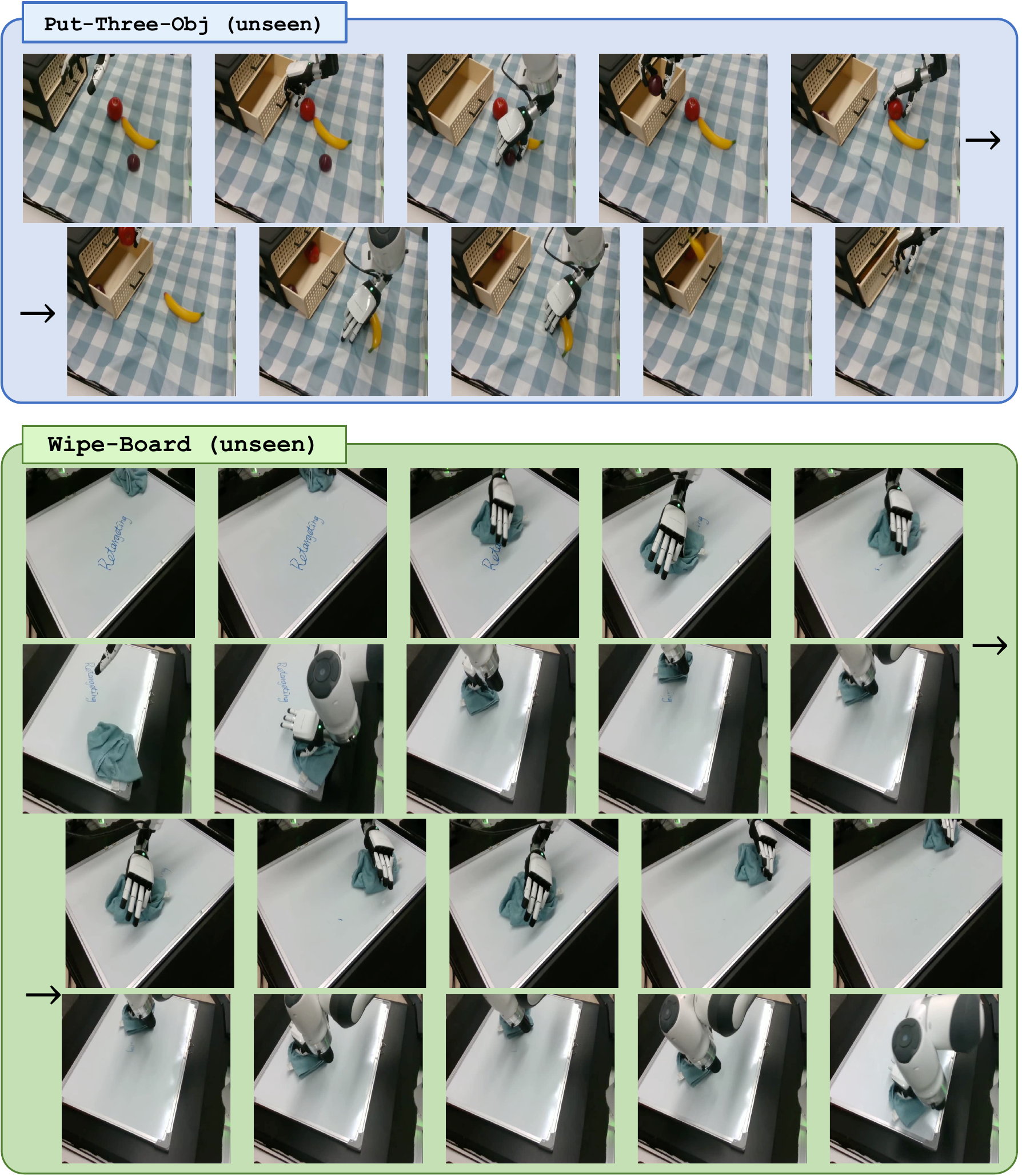}
    \caption{
    \textbf{Successful real-robot rollouts (unseen setting).}
    \texttt{Put-Three-Obj}: the policy corrects an initial misalignment under altered tablecloth texture.
    \texttt{Wipe-Board}: the policy adaptively revisits residual ink under changed marker color.
    }
    \label{fig:success_unseen}
\end{figure}

\begin{figure}[h]
    \centering
    \includegraphics[width=0.9\linewidth]{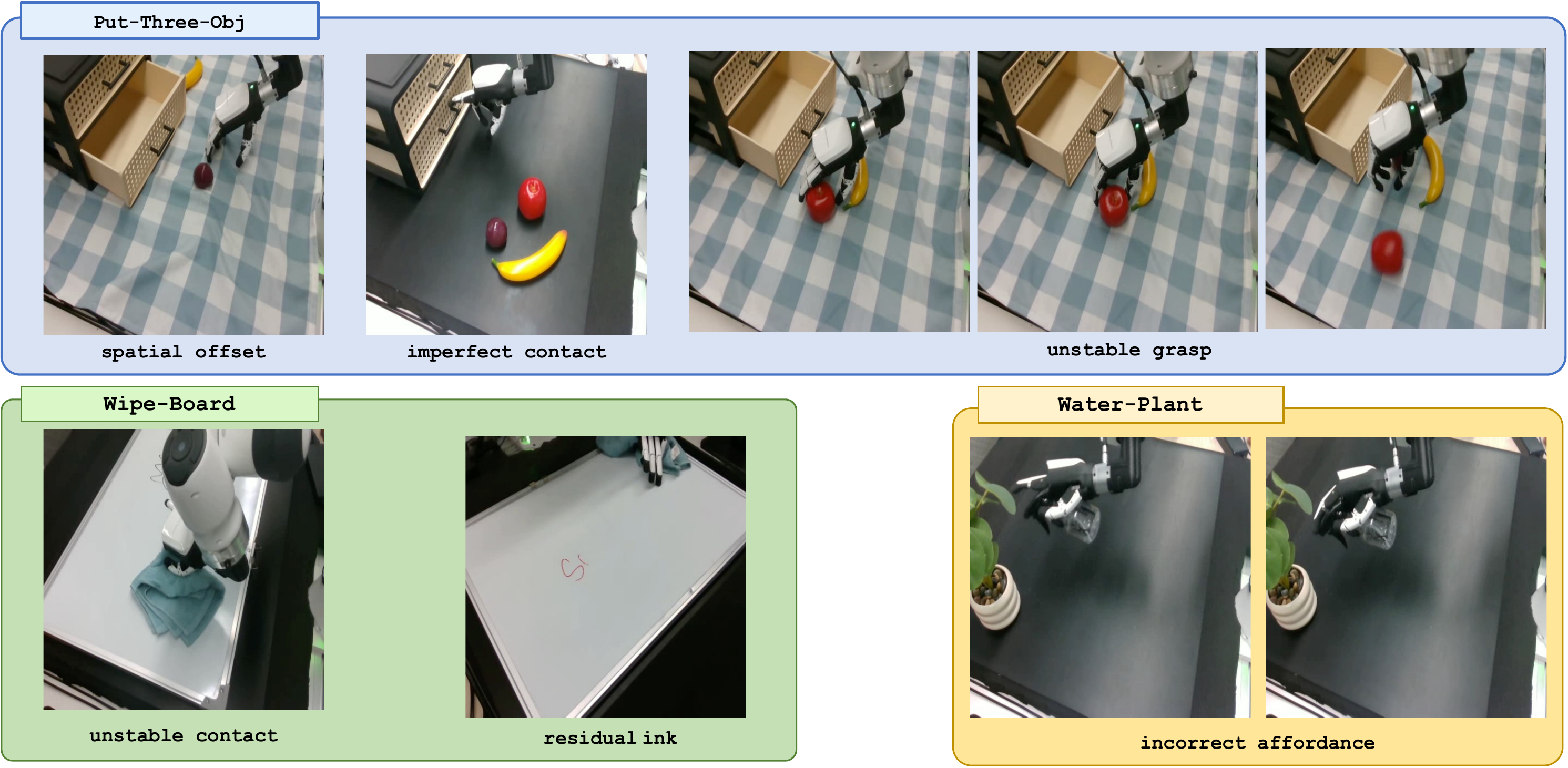}
    \caption{
    \textbf{Failure cases across real-robot tasks}.
    \texttt{Put-Three-Obj}: spatial misalignment, incomplete contact, and unstable grasp.
    \texttt{Wipe-Board}: insufficient planar contact and persistent residual ink.
    \texttt{Water-Plant}: incorrect affordance reasoning leading to wrong bottle orientation.
    }
    \label{fig:bad_cases}
\end{figure}

\noindent\textbf{Robot system.}
We deploy JALA on a real-world platform consisting of a 7-DoF Franka Research~3 arm and a 6-DoF Inspire dexterous hand.
The Inspire hand provides independent actuation of the thumb, index, middle, ring, and little fingers plus a lateral thumb rotation, enabling a rich repertoire of power and precision grasps.
Third-person RGB observations are captured by Intel RealSense~L515 depth cameras mounted at fixed positions.
The policy receives these RGB frames together with the robot's proprioceptive state (joint positions and gripper configuration) and outputs action chunks to execute.

\noindent\textbf{Task design.}
We design three multi-step manipulation tasks that probe complementary aspects of dexterous control (Fig.~\ref{fig:setup}).
(1) \textbf{Put-Three-Obj} requires the robot to open a drawer, place three fruits inside, and close the drawer, exercising multi-object localization, sequential pick-and-place, and long-horizon coordination across five ordered subtasks.
(2) \textbf{Wipe-Board} asks the robot to grasp a cloth, wipe pen marks from a whiteboard, and remove visible ink, requiring sustained planar contact, region-aware perception under occlusion, and iterative wiping across three subtasks.
(3) \textbf{Water-Plant} requires the robot to grasp a spray bottle, reposition it toward a plant, and activate the trigger, testing fine-grained multi-finger coordination and spatial reasoning across three subtasks.
To reflect the sensing constraints of each task, Put-Three-Obj and Water-Plant rely on a single fixed camera view, whereas Wipe-Board uses two views to mitigate occlusion caused by the arm during wiping.

\noindent\textbf{Data collection and evaluation protocol.}
For each task, we collect 50 teleoperated demonstrations via a human operator controlling the arm-hand system. Each demonstration is paired with third-person RGB observations and a textual task instruction for downstream post-training (Fig.~\ref{fig:teleop_examples}).
Evaluation is performed over 10 rollouts per task, and we report the average proportion of completed subtasks (\textbf{Completion Ratio}) as the success metric.
To assess generalization under visual distribution shifts, we further introduce unseen variations for two of the three tasks. In Put-Three-Obj, the tablecloth pattern is changed to a previously unseen texture. In Wipe-Board, the marker color is altered from the training distribution (Fig.~\ref{fig:setup}).
Water-Plant does not include an unseen variant because its visual background is already relatively simple and stable.

\input{tables/real}

\noindent\textbf{Quantitative results.}
Across all three real-world manipulation tasks (Table~\ref{table:real_robot}), JALA consistently outperforms both strong baselines and our own ablations, with \textbf{JALA-dino} achieving the best completion rates in every task. The advantage is consistent across both seen and unseen settings, rather than coming from a single favorable scenario. In particular, JALA shows clear gains on multi-step object placement and surface-interaction tasks, and it also maintains a strong lead on \textit{Water-Plant}. These results indicate that the joint-aligned latent action space learned from human videos transfers effectively to real-world 3D manipulation.

\noindent\textbf{Robustness to visual shifts.} A particularly important finding is JALA's robustness under visual and environmental shifts. In the unseen variants of \textit{Put-Three-Obj} and \textit{Wipe-Board}, where the tablecloth or marker color changes, baseline performance degrades sharply, whereas JALA-dino remains much more stable (e.g., only a small drop from 60.0\% to 58.0\% on \textit{Put-Three-Obj}). This pattern suggests that JALA relies less on superficial appearance cues and more on action-relevant dynamics and interaction structure. The ablation results further support this interpretation: \textit{JALA w/o align} consistently underperforms full JALA, especially in unseen settings, indicating that alignment to latent actions is a key ingredient for learning representations that remain effective under domain shifts and physical variability. Overall, the real-robot experiments demonstrate that JALA improves not only task success in multi-step manipulation, but also robustness and transferability across changing environments.

\noindent\textbf{Successful rollout analysis.}
We qualitatively analyze successful rollouts to understand the behavioral competence learned by JALA (Figs.~\ref{fig:success_seen} and~\ref{fig:success_unseen}).
In Put-Three-Obj, the policy demonstrates accurate spatial reasoning, precise pre-grasp alignment, and consistent grasp stability across objects of different shapes.
A noteworthy behavior emerges in the unseen case. While attempting to grasp the banana, the initial approach is slightly misaligned due to its curved geometry and the shifted tablecloth texture.
Instead of committing to an incorrect lift, the policy retracts, re-positions its wrist, and performs a corrected grasp on the second attempt.
This self-corrective motion is not explicitly supervised but arises from the latent action modeling, indicating that the policy has internalized a feedback-driven recovery strategy.
In Wipe-Board, the ink is rarely removed in a single stroke. The policy repeatedly reorients the cloth and revisits regions where residual ink remains, reflecting closed-loop adaptation rather than a fixed wiping trajectory. This iterative refinement behavior persists even when marker colors change in the unseen setting, suggesting strong generalization to new visual distributions.
In Water-Plant, the policy consistently exhibits a reliable grasp of the spray bottle and a stable reorientation toward the plant. Trigger activation, which requires precise multi-finger coordination, is executed cleanly and without premature release.

\noindent\textbf{Failure mode analysis.}
We further analyze representative failure cases to identify remaining limitations (Fig.~\ref{fig:bad_cases}).
In Put-Three-Obj, failures arise from compounding perception-control mismatches. Subtle spatial offsets in the predicted pre-grasp pose lead the fingers to contact the fruit at an unintended angle. Even when initial contact is made, insufficient finger closure or small wrist-orientation errors cause the fruit to slip during transport, especially for the elongated banana whose geometry amplifies torque instability.
In Wipe-Board, two primary failure factors are observed. First, the cloth is sometimes not pressed firmly or flatly enough against the board surface, causing it to fold or lose planar contact. Second, the model occasionally fails to revisit lightly marked regions, suggesting that the closed-loop perception sometimes underestimates faint ink traces.
In Water-Plant, most failures arise from incorrect affordance reasoning. The policy grasps the spray bottle but positions it at an incorrect orientation relative to the plant or the trigger mechanism, failing to align the nozzle toward the target or the index-finger actuation direction with the trigger.
These cases highlight remaining challenges in precise alignment, contact stability, and fine-grained affordance modeling.

%% file: tables/motion_gen.tex
\begin{table}[t]
\centering
\caption{
Comparison of hand motion generation and prediction tasks on both Lab and Wild splits.
}
\label{table:hand_motion_generation}
%\vspace{-2mm}
\setlength{\tabcolsep}{4pt}{
\scalebox{0.85}{
\begin{tabular}{m{3.5cm} *{8}{>{\centering\arraybackslash}m{1.2cm}}}
\toprule
\multirow{2}{*}{\textbf{Model}}
& \multicolumn{2}{c}{\textbf{MPJPE} $\downarrow$} 
& \multicolumn{2}{c}{\textbf{PA-MPJPE} $\downarrow$}
& \multicolumn{2}{c}{\textbf{MWTE} $\downarrow$} 
& \multicolumn{2}{c}{\textbf{MDE} $\downarrow$}
\\
\cmidrule(lr){2-3}
\cmidrule(lr){4-5}
\cmidrule(lr){6-7}
\cmidrule(lr){8-9}
& Lab & Wild
& Lab & Wild
& Lab & Wild
& Lab & Wild
\\
 \midrule \addlinespace 
\rowcolor{BlockA!30}
\multicolumn{9}{l}{\textbf{\# Next Token Prediction}} \\
\rowcolor{BlockA!30}
Being-H0        & 7.61 & 16.91 & 1.34 & 3.81 & 6.03 & 14.54 & 7.16 & 18.33 \\
\rowcolor{BlockA!30}
Being-H0+dino   & 7.54 & 15.14 & 0.90 & 2.78 & 5.85 & 13.65 & 6.95 & 17.17 \\
\addlinespace
\rowcolor{BlockB!30}
\multicolumn{9}{l}{\textbf{\# Masked Chunk Prediction}} \\
% LatentAct & & & & & & & & & &  \\
\rowcolor{BlockB!30}
JALA w/o align      & 7.72 & 15.73 & \textbf{0.89 }& 2.34 & 6.18 & 14.02 & 8.09 & 16.23 \\
\rowcolor{BlockB!30}
JALA w/o latent     & 8.26 & 20.34 & 1.83 & 3.94 & 8.02 & 17.13 & 10.17 & 26.74 \\
\rowcolor{BlockB!30}
JALA-dino           & 7.16 & \textbf{11.02} & 0.91 & \textbf{1.12} & \textbf{5.77} &\textbf{ 9.79 } & 7.24 & \textbf{11.04} \\
\rowcolor{BlockB!30}
JALA-vjepa          & \textbf{7.05} & 11.54 & 0.94 & 1.32 & 5.85 & 10.04 & \textbf{6.73} & 11.87 \\
\bottomrule
\end{tabular}
}}
%\vspace{-3mm}
\end{table}

%% file: tables/libero_2.tex
\begin{table}[t]
\centering
\caption{
Two-view LIBERO results showing success rates (\%) across task categories and overall average.
\colorbox{BlockA!30}{Blue rows} denote models trained on the action-available subset only, while \colorbox{BlockB!30}{green rows} denote models pretrained on the full UniHand-Mix.
LAPA$^\dagger$ is LAPA retrained on our data with the JALA backbone.
}
\label{table:libero_two_view}
\setlength{\tabcolsep}{4pt}{
\scalebox{0.9}{
\begin{tabular}{m{3.0cm} *{5}{>{\centering\arraybackslash}m{1.2cm}}}
\toprule
\textbf{Model} & \textbf{Spatial} & \textbf{Object} & \textbf{Goal} & \textbf{Long} & \textbf{Average} \\
\midrule
LAPA~\citep{ye2024lapa}               & 83.4 & 87.6 & 78.2 & 68.8 & 79.5 \\
MolmoAct~\citep{lee2025molmoact}      & 87.0 & 95.4 & 87.6 & 77.2 & 86.6 \\
$\pi$0-FAST~\citep{pertsch2025fast}   & 96.4 & 96.8 & 88.6 & 60.2 & 85.5 \\
GR00T N1.5~\citep{bjorck2025gr00tn1}  & 94.4 & 97.6 & 93.0 & 90.6 & 93.9 \\
$\pi$0~\citep{black2024pi_0}          & \textbf{96.8} & \textbf{98.8} & 95.8 & 85.2 & 94.2 \\
UniVLA~\citep{univla2024}             & 95.4 & \textbf{98.8} & 93.6 & 94.0 & 95.5 \\
\rowcolor{BlockA!30}
Being-H0~\citep{beingbeyond2025beingh0} & 92.6 & 96.8 & 94.0 & 77.4 & 90.2 \\
\rowcolor{BlockA!30}
JALA-act                               & 93.4 & 97.8 & 94.2 & 91.8 & 94.3 \\
\rowcolor{BlockA!30}
JALA w/o dec.                          & 64.6 & 58.4 & 61.2 & 42.2 & 56.6 \\
\rowcolor{BlockB!30}
LAPA$^\dagger$                         & 87.4 & 91.2 & 90.0 & 65.4 & 83.5 \\
\rowcolor{BlockB!30}
JALA$^\star$                           & 95.2 & 96.4 & 97.2 & 94.0 & 95.7 \\
\rowcolor{BlockB!30}
JALA-vjepa                             & 95.4 & 98.0 & \textbf{98.0} & 94.8 & 96.6 \\
\rowcolor{BlockB!30}
JALA-dino                              & 96.0 & 98.2 & 97.4 & \textbf{96.0} & \textbf{96.9} \\
\bottomrule
\end{tabular}
}}
\end{table}

%% file: tables/libero.tex
\begin{table}[t]
\centering
\caption{
Single-view LIBERO results showing success rates (\%) across task categories and overall average.
Models are grouped by size: $>$3B parameters (top) and $\le$3B parameters (bottom). 
The results of LAPA$^\star$ are reported in UniVLA.
UniVLA-human$^\dagger$ uses only human pretraining data. 
UniVLA-full$^{\dagger\dagger}$ incorporates Bridge-V2~\citep{walke2023bridgedata} for pretraining.
}
\label{table:libero}
%\vspace{-2mm}
\setlength{\tabcolsep}{4pt}{
\scalebox{0.9}{
\begin{tabular}{m{3.2cm} *{5}{>{\centering\arraybackslash}m{1.2cm}}}
\toprule
\textbf{Model} & \textbf{Spatial} & \textbf{Object} & \textbf{Goal} & \textbf{Long} & \textbf{Average} \\
\midrule \addlinespace
\rowcolor{BlockA!30}
\multicolumn{6}{l}{\textbf{\# $>$ 3B Backbones}} \\
\rowcolor{BlockA!30}
LAPA$^\star$~\citep{ye2024lapa}   & 73.8 & 74.6 & 58.8 & 55.4 & 65.7 \\
\rowcolor{BlockA!30}
OpenVLA~\citep{kim2025openvla}             & 84.7 & 88.4 & 79.2 & 53.7 & 76.5 \\
\rowcolor{BlockA!30}
TriVLA~\citep{liu2025trivla}              & 91.2 & 93.8 & 89.8 & 73.2 & 87.0 \\
\rowcolor{BlockA!30}
4D-VLA~\citep{zhang20254d}              & 88.9 & 95.2 & 90.9 & 79.1 & 88.5 \\
\rowcolor{BlockA!30}
UniVLA-human$^\dagger$~\citep{univla2024}      & 91.2 & 94.2 & 90.2 & 79.4 & 88.7 \\
\rowcolor{BlockA!30}
UniVLA-full$^{\dagger\dagger}$~\citep{univla2024}         & \textbf{96.5} & \textbf{96.8} & \textbf{95.6} & \textbf{92.0} & \textbf{95.2} \\
\addlinespace
\rowcolor{BlockB!30}
\multicolumn{6}{l}{\textbf{\# $\le$ 3B Backbones}} \\
\rowcolor{BlockB!30}
Diffusion Policy~\citep{chi2023diffusion}    & 78.3 & 92.5 & 68.3 & 50.5 & 72.4 \\
\rowcolor{BlockB!30}
Octo~\citep{team2024octo}                & 78.9 & 85.7 & 84.6 & 51.1 & 75.1 \\
\rowcolor{BlockB!30}
UniACT~\citep{zhang2025uniact}              & 77.0 & 87.0 & 77.0 & 70.0 & 77.8 \\
\rowcolor{BlockB!30}
SpatialVLA~\citep{qu2025spatialvla}          & 88.2 & 89.9 & 78.6 & 55.5 & 78.1 \\
\rowcolor{BlockB!30}
DiT Policy~\citep{hou2024diffusion}          & 84.2 & 96.3 & 85.4 & 63.8 & 82.4 \\
\rowcolor{BlockB!30}
ThinkAct~\citep{huang2025thinkact}            & 88.3 & 91.4 & 87.1 & 70.9 & 84.4 \\
\rowcolor{BlockB!30}
Being-H0~\citep{beingbeyond2025beingh0}         & 86.6 & 92.8 & 89.6 & 70.4 & 84.9 \\
\rowcolor{BlockB!30}
GR00T N1.5~\citep{bjorck2025gr00tn1}          & 91.4 & 97.6 & 94.0 & 85.6 & 92.1 \\
\rowcolor{BlockB!30}
JALA w/o align      & 89.4 & 91.2 & 90.0 & 72.6 & 85.8 \\
\rowcolor{BlockB!30}
JALA w/o latent      & 80.4 & 83.6 & 75.2 & 68.6 & 77.0 \\
\rowcolor{BlockB!30}
JALA-vjepa          & \textbf{91.6} & \textbf{98.2} & 94.4 & 84.2 & 92.1 \\
\rowcolor{BlockB!30}
JALA-dino           & 90.4 & 96.4 & \textbf{95.2} & \textbf{87.2} & \textbf{92.3} \\
\bottomrule
\end{tabular}
}}
%\vspace{-3mm}
\end{table}

%% file: tables/robocasa.tex
\begin{table}[t]
\centering
\caption{
Results on the RoboCasa benchmark and GR1 tabletop tasks.
We report success rates (\%).
\colorbox{BlockA!30}{Blue rows} denote models trained on the action-available subset only, while \colorbox{BlockB!30}{green rows} denote models pretrained on the full UniHand-Mix.
}
\label{table:robocasa}
\setlength{\tabcolsep}{6pt}{
\scalebox{0.9}{
\begin{tabular}{lccc}
\toprule
\textbf{Model} & \textbf{RoboCasa Syn.} & \textbf{RoboCasa Human} & \textbf{GR1 Tabletop} \\
\midrule
GR00T N1.5        & 20.83 & 35.17 & 20.41 \\
LAPA              & 16.25 & 22.42 & 11.42 \\
\rowcolor{BlockA!30}
Being-H0          & 23.83 & 31.33 & 12.91 \\
\rowcolor{BlockA!30}
JALA-act          & 24.92 & 32.42 & 20.25 \\
\rowcolor{BlockA!30}
JALA w/o dec.     & 14.25 & 19.33 & 9.25 \\
\rowcolor{BlockB!30}
LAPA$^\dagger$    & 20.25 & 27.33 & 13.50 \\
\rowcolor{BlockB!30}
JALA$^\star$      & 25.33 & 33.83 & 24.50 \\
\rowcolor{BlockB!30}
JALA              & \textbf{27.58} & \textbf{35.42} & \textbf{26.33} \\
\bottomrule
\end{tabular}
}}
\end{table}

%% file: tables/real.tex
\begin{table}[h]
\centering
\caption{
Real-world robot performance measured by average subtask completion rate (\%) on three multi-step manipulation tasks. 
Each policy is evaluated over 10 rollouts per task. 
}
\label{table:real_robot}
\setlength{\tabcolsep}{6pt}{
\scalebox{0.9}{
\begin{tabular}{m{3.0cm} *{4}{>{\centering\arraybackslash}m{1.2cm}} >{\centering\arraybackslash}m{2.2cm}}
\toprule
\multirow{2}{*}{\textbf{Model}} & 
\multicolumn{2}{c}{\textbf{Put-Three-Obj}} &
\multicolumn{2}{c}{\textbf{Wipe-Board}} &
\textbf{Water-Plant} \\
\cmidrule(lr){2-3} \cmidrule(lr){4-5} \cmidrule(lr){6-6}
 & Seen & Unseen & Seen & Unseen & Seen  \\
\midrule
Being-H0        & 38.0 & 16.0 & 40.0 & 33.3 & 36.7  \\
GR00T N1.5      & 48.0 & 28.0 & 56.7 & 43.3 & 53.3  \\
JALA w/o align  & 40.0 & 32.0 & 53.3 & 43.3 & 56.7  \\
JALA-vjepa      & 38.0 & 34.0 & 66.7 & 60.0 & 66.7  \\
JALA-dino       & \textbf{60.0} & \textbf{58.0} & \textbf{83.3} & \textbf{80.0} & \textbf{73.3}\\
\bottomrule
\end{tabular}
}}
\end{table}

%% file: sections_arxiv/07_conclusion.tex
\section{Conclusion}
In this work, we presented \textbf{JALA}, a pretraining framework that rethinks how to represent latent actions for scalable VLA pretraining with human videos. By introducing the joint alignment between predictive embeddings from the VLA context and IDM-derived latent actions, JALA moves beyond reconstruction-heavy paradigms and instead captures motion cues without detailed distortion. This design enables a unified latent action space that works consistently across lab-annotated and in-the-wild human data. To support pretraining at scale, we constructed \textbf{UniHand-Mix}, a 7.5M-sample dataset combining over 2,000 hours of annotated and unconstrained human videos. This hybrid resource provides both reliable physical anchors and broad task diversity, and our experiments demonstrate that JALA can effectively leverage both sources. Results show great improvements in hand motion generation and gains on downstream robot manipulation benchmarks, surpassing existing VLA methods with comparable or larger model sizes.  Overall, our study highlights predictive embeddings as a practical and scalable route for bridging human video data with robotic learning. We believe this perspective opens opportunities for learning from even larger and more diverse human corpora.  

%% file: main_arxiv.bbl
\begin{thebibliography}{10}

\bibitem{zitkovich2023rt}
Brianna Zitkovich, Tianhe Yu, Sichun Xu, Peng Xu, Ted Xiao, Fei Xia, Jialin Wu, Paul Wohlhart, Stefan Welker, Ayzaan Wahid, et~al.
\newblock Rt-2: Vision-language-action models transfer web knowledge to robotic control.
\newblock In {\em Proc. of Conference on Robot Learning}, 2023.

\bibitem{black2024pi_0}
Kevin Black, Noah Brown, Danny Driess, Adnan Esmail, Michael Equi, Chelsea Finn, Niccolo Fusai, Lachy Groom, Karol Hausman, Brian Ichter, et~al.
\newblock {\(\pi\)}\({}_{\mbox{0}}\): A vision-language-action flow model for general robot control.
\newblock {\em arXiv preprint arXiv:2410.24164}, 2024.

\bibitem{bjorck2025gr00tn1}
Johan Bjorck, Fernando Casta{\~{n}}eda, Nikita Cherniadev, Xingye Da, Runyu Ding, Linxi Fan, Yu~Fang, Dieter Fox, Fengyuan Hu, Spencer Huang, et~al.
\newblock {GR00T N1}: An open foundation model for generalist humanoid robots.
\newblock {\em arXiv preprint arXiv:2503.14734}, 2025.

\bibitem{o2024open}
Abby O'Neill, Abdul Rehman, Abhiram Maddukuri, Abhishek Gupta, Abhishek Padalkar, Abraham Lee, Acorn Pooley, Agrim Gupta, Ajay Mandlekar, Ajinkya Jain, et~al.
\newblock {Open X-Embodiment}: Robotic learning datasets and {RT-X} models.
\newblock In {\em Proc. of International Conference on Robotics and Automation}, pages 6892--6903, 2024.

\bibitem{khazatsky2024droid}
Alexander Khazatsky, Karl Pertsch, Suraj Nair, Ashwin Balakrishna, Sudeep Dasari, Siddharth Karamcheti, Soroush Nasiriany, Mohan~Kumar Srirama, Lawrence~Yunliang Chen, Kirsty Ellis, et~al.
\newblock {DROID}: A large-scale in-the-wild robot manipulation dataset.
\newblock {\em arXiv preprint arXiv:2403.12945}, 2024.

\bibitem{touvron2023llama}
Hugo Touvron, Thibaut Lavril, Gautier Izacard, Xavier Martinet, Marie-Anne Lachaux, Timoth{\'e}e Lacroix, Baptiste Rozi{\`e}re, Naman Goyal, Eric Hambro, Faisal Azhar, et~al.
\newblock Llama: Open and efficient foundation language models.
\newblock {\em arXiv preprint arXiv:2302.13971}, 2023.

\bibitem{li2024llava}
Bo~Li, Yuanhan Zhang, Dong Guo, Renrui Zhang, Feng Li, Hao Zhang, Kaichen Zhang, Peiyuan Zhang, Yanwei Li, Ziwei Liu, et~al.
\newblock Llava-onevision: Easy visual task transfer.
\newblock {\em arXiv preprint arXiv:2408.03326}, 2024.

\bibitem{kim2025openvla}
Moo~Jin Kim, Karl Pertsch, Siddharth Karamcheti, Ted Xiao, Ashwin Balakrishna, Suraj Nair, Rafael Rafailov, Ethan~P Foster, Pannag~R Sanketi, Quan Vuong, et~al.
\newblock Openvla: An open-source vision-language-action model.
\newblock In {\em Proc. of Conference on Robot Learning}, 2025.

\bibitem{team2024octo}
{Octo Model Team}, Dibya Ghosh, Homer Walke, Karl Pertsch, Kevin Black, Oier Mees, Sudeep Dasari, Joey Hejna, Charles Xu, Jianlan Luo, et~al.
\newblock {Octo}: An open-source generalist robot policy.
\newblock In {\em Proc. of Robotics: Science and Systems}, 2024.

\bibitem{beingbeyond2025beingh0}
Hao Luo, Yicheng Feng, Wanpeng Zhang, Sipeng Zheng, Ye~Wang, Haoqi Yuan, Jiazheng Liu, Chaoyi Xu, Qin Jin, and Zongqing Lu.
\newblock Being-h0: Vision-language-action pretraining from large-scale human videos.
\newblock {\em arXiv preprint arXiv:2507.15597}, 2025.

\bibitem{yang2025egovla}
Ruihan Yang, Qinxi Yu, Yecheng Wu, Rui Yan, Borui Li, An-Chieh Cheng, Xueyan Zou, Yunhao Fang, Hongxu Yin, Sifei Liu, Song Han, Yao Lu, and Xiaolong Wang.
\newblock Egovla: Learning vision-language-action models from egocentric human videos.
\newblock {\em arXiv preprint arXiv:2507.2507.12440}, 2025.

\bibitem{fan2023arctic}
Zicong Fan, Omid Taheri, Dimitrios Tzionas, Muhammed Kocabas, Manuel Kaufmann, Michael~J Black, and Otmar Hilliges.
\newblock {ARCTIC}: A dataset for dexterous bimanual hand-object manipulation.
\newblock In {\em Proc. of Computer Vision and Pattern Recognition}, 2023.

\bibitem{chao2021dexycb}
Yu-Wei Chao, Wei Yang, Yu~Xiang, Pavlo Molchanov, Ankur Handa, Jonathan Tremblay, Yashraj~S Narang, Karl Van~Wyk, Umar Iqbal, Stan Birchfield, et~al.
\newblock {DexYCB}: A benchmark for capturing hand grasping of objects.
\newblock In {\em Proc. of Computer Vision and Pattern Recognition}, 2021.

\bibitem{egodex}
Ryan Hoque, Peide Huang, David~J. Yoon, Mouli Sivapurapu, and Jian Zhang.
\newblock Egodex: Learning dexterous manipulation from large-scale egocentric video.
\newblock {\em arXiv preprint arXiv:2505.11709}, 2025.

\bibitem{liu2022hoi4d}
Yunze Liu, Yun Liu, Che Jiang, Kangbo Lyu, Weikang Wan, Hao Shen, Boqiang Liang, Zhoujie Fu, He~Wang, and Li~Yi.
\newblock Hoi4d: A 4d egocentric dataset for category-level human-object interaction.
\newblock In {\em Proc. of Computer Vision and Pattern Recognition}, 2022.

\bibitem{grauman2022ego4d}
Kristen Grauman, Andrew Westbury, Eugene Byrne, Zachary Chavis, Antonino Furnari, Rohit Girdhar, Jackson Hamburger, Hao Jiang, Miao Liu, Xingyu Liu, et~al.
\newblock {Ego4D}: Around the world in 3,000 hours of egocentric video.
\newblock In {\em Proc. of Computer Vision and Pattern Recognition}, pages 18995--19012, 2022.

\bibitem{perrett2025hd}
Toby Perrett, Ahmad Darkhalil, Saptarshi Sinha, Omar Emara, Sam Pollard, Kranti~Kumar Parida, Kaiting Liu, Prajwal Gatti, Siddhant Bansal, Kevin Flanagan, et~al.
\newblock Hd-epic: A highly-detailed egocentric video dataset.
\newblock In {\em Proc. of Computer Vision and Pattern Recognition}, 2025.

\bibitem{damen2022rescaling}
Dima Damen, Hazel Doughty, Giovanni~Maria Farinella, Antonino Furnari, Jian Ma, Evangelos Kazakos, Davide Moltisanti, Jonathan Munro, Toby Perrett, Will Price, et~al.
\newblock Rescaling egocentric vision: Collection, pipeline and challenges for {EPIC-KITCHENS-100}.
\newblock {\em International Journal of Computer Vision}, 130:33--55, 2022.

\bibitem{ye2024lapa}
Seonghyeon Ye, Joel Jang, Byeongguk Jeon, Sejune Joo, Jianwei Yang, Baolin Peng, Ajay Mandlekar, Reuben Tan, Yu-Wei Chao, Bill~Yuchen Lin, et~al.
\newblock Latent action pretraining from videos.
\newblock In {\em International Journal of Computer Vision}, 2025.

\bibitem{nair2023r3m}
Suraj Nair, Aravind Rajeswaran, Vikash Kumar, Chelsea Finn, and Abhinav Gupta.
\newblock R3m: A universal visual representation for robot manipulation.
\newblock In {\em Proc. of Conference on Robot Learning}, 2023.

\bibitem{radosavovic2023real}
Ilija Radosavovic, Tete Xiao, Stephen James, Pieter Abbeel, Jitendra Malik, and Trevor Darrell.
\newblock Real-world robot learning with masked visual pre-training.
\newblock In {\em Proc. of Conference on Robot Learning}, 2023.

\bibitem{wu2024gr1}
Hongtao Wu, Ya~Jing, Chilam Cheang, Guangzeng Chen, Jiafeng Xu, Xinghang Li, Minghuan Liu, Hang Li, and Tao Kong.
\newblock Unleashing large-scale video generative pre-training for visual robot manipulation.
\newblock In {\em International Journal of Computer Vision}, 2024.

\bibitem{cheang2024gr2}
Chilam Cheang, Guangzeng Chen, Ya~Jing, Tao Kong, Hang Li, Yifeng Li, Yuxiao Liu, Hongtao Wu, Jiafeng Xu, Yichu Yang, et~al.
\newblock {GR-2}: A generative video-language-action model with web-scale knowledge for robot manipulation.
\newblock {\em arXiv preprint arXiv:2410.06158}, 2024.

\bibitem{video2skill2024}
Video2Skill Team.
\newblock Video2skill: Adapting foundation models for task-level robot manipulation.
\newblock {\em arXiv preprint}, 2024.

\bibitem{chen2024igor}
Xiaoyu Chen, Junliang Guo, Tianyu He, Chuheng Zhang, Pushi Zhang, Derek~Cathera Yang, Li~Zhao, and Jiang Bian.
\newblock {IGOR}: Image-goal representations are the atomic control units for foundation models in embodied ai.
\newblock {\em arXiv preprint arXiv:2411.00785}, 2024.

\bibitem{lipman2022flow}
Yaron Lipman, Ricky~TQ Chen, Heli Ben-Hamu, Maximilian Nickel, and Matt Le.
\newblock Flow matching for generative modeling.
\newblock {\em arXiv preprint arXiv:2210.02747}, 2022.

\bibitem{liu2023libero}
Bo~Liu, Yifeng Zhu, Chongkai Gao, Yihao Feng, Qiang Liu, Yuke Zhu, and Peter Stone.
\newblock Libero: Benchmarking knowledge transfer for lifelong robot learning.
\newblock {\em arXiv preprint arXiv:2306.03310}, 2023.

\bibitem{robocasa2024}
Soroush Nasiriany, Abhiram Maddukuri, Lance Zhang, Adeet Parikh, Aaron Lo, Abhishek Joshi, Ajay Mandlekar, and Yuke Zhu.
\newblock Robocasa: Large-scale simulation of everyday tasks for generalist robots.
\newblock In {\em Proc. of Robotics: Science and Systems}, 2024.

\bibitem{open2024open}
Abhishek Padalkar and et\ al.
\newblock Open x-embodiment: Robotic learning datasets and rt-x models.
\newblock {\em arXiv preprint arXiv:2310.08864}, 2023.

\bibitem{bu2025agibot}
Qingwen Bu, Jisong Cai, Li~Chen, Xiuqi Cui, Yan Ding, Siyuan Feng, Shenyuan Gao, Xindong He, Xuan Hu, Xu~Huang, et~al.
\newblock Agibot world colosseo: A large-scale manipulation platform for scalable and intelligent embodied systems.
\newblock {\em arXiv preprint arXiv:2503.06669}, 2025.

\bibitem{brohan2022rt1}
Anthony Brohan, Noah Brown, Justice Carbajal, Yevgen Chebotar, Joseph Dabis, Chelsea Finn, Keerthana Gopalakrishnan, Karol Hausman, Alex Herzog, Jasmine Hsu, et~al.
\newblock {RT-1}: Robotics transformer for real-world control at scale.
\newblock In {\em Proc. of Robotics: Science and Systems}, 2023.

\bibitem{pertsch2025fast}
Karl Pertsch, Kyle Stachowicz, Brian Ichter, Danny Driess, Suraj Nair, Quan Vuong, Oier Mees, Chelsea Finn, and Sergey Levine.
\newblock {FAST}: Efficient action tokenization for vision-language-action models.
\newblock {\em arXiv preprint arXiv:2501.09747}, 2025.

\bibitem{zhong2025survey}
Yifan Zhong, Fengshuo Bai, Shaofei Cai, Xuchuan Huang, Zhang Chen, Xiaowei Zhang, Yuanfei Wang, Shaoyang Guo, Tianrui Guan, Ka~Nam Lui, et~al.
\newblock A survey on vision-language-action models: An action tokenization perspective.
\newblock {\em arXiv preprint arXiv:2507.01925}, 2025.

\bibitem{intelligence2025pi_}
Physical Intelligence, Kevin Black, Noah Brown, James Darpinian, Karan Dhabalia, Danny Driess, Adnan Esmail, Michael Equi, Chelsea Finn, Niccolo Fusai, et~al.
\newblock \(\pi_ {0.5}\): a vision-language-action model with open-world generalization.
\newblock {\em arXiv preprint arXiv:2504.16054}, 2025.

\bibitem{driess2025knowledge}
Danny Driess, Jost~Tobias Springenberg, Brian Ichter, Lili Yu, Adrian Li-Bell, Karl Pertsch, Allen~Z Ren, Homer Walke, Quan Vuong, Lucy~Xiaoyang Shi, et~al.
\newblock Knowledge insulating vision-language-action models: Train fast, run fast, generalize better.
\newblock {\em arXiv preprint arXiv:2505.23705}, 2025.

\bibitem{liu2024rdt}
Songming Liu, Lingxuan Wu, Bangguo Li, Hengkai Tan, Huayu Chen, Zhengyi Wang, Ke~Xu, Hang Su, and Jun Zhu.
\newblock Rdt-1b: a diffusion foundation model for bimanual manipulation.
\newblock {\em arXiv preprint arXiv:2410.07864}, 2024.

\bibitem{wen2025dexvla}
Junjie Wen, Yichen Zhu, Jinming Li, Zhibin Tang, Chaomin Shen, and Feifei Feng.
\newblock Dexvla: Vision-language model with plug-in diffusion expert for general robot control.
\newblock {\em arXiv preprint arXiv:2502.05855}, 2025.

\bibitem{shukor2025smolvla}
Mustafa Shukor, Dana Aubakirova, Francesco Capuano, Pepijn Kooijmans, Steven Palma, Adil Zouitine, Michel Aractingi, Caroline Pascal, Martino Russi, Andres Marafioti, et~al.
\newblock Smolvla: A vision-language-action model for affordable and efficient robotics.
\newblock {\em arXiv preprint arXiv:2506.01844}, 2025.

\bibitem{univla2024}
UniVLA Team.
\newblock {UniVLA}: Unified vision-language-action model for cross-embodiment robotic learning.
\newblock {\em arXiv preprint}, 2024.

\bibitem{banerjee2025hot3d}
Prithviraj Banerjee, Sindi Shkodrani, Pierre Moulon, Shreyas Hampali, Shangchen Han, Fan Zhang, Linguang Zhang, Jade Fountain, Edward Miller, Selen Basol, et~al.
\newblock Hot3d: Hand and object tracking in 3d from egocentric multi-view videos.
\newblock In {\em Proc. of Computer Vision and Pattern Recognition}, 2025.

\bibitem{singh2025deep}
Himanshu~Gaurav Singh, Pieter Abbeel, Jitendra Malik, and Antonio Loquercio.
\newblock Deep sensorimotor control by imitating predictive models of human motion.
\newblock {\em arXiv preprint arXiv:2508.18691}, 2025.

\bibitem{yang2015robot}
Yezhou Yang, Yi~Li, Cornelia Fermuller, and Yiannis Aloimonos.
\newblock Robot learning manipulation action plans by" watching" unconstrained videos from the world wide web.
\newblock In {\em The Association for the Advancement of Artificial Intelligence}, 2015.

\bibitem{seo2022reinforcement}
Younggyo Seo, Kimin Lee, Stephen~L James, and Pieter Abbeel.
\newblock Reinforcement learning with action-free pre-training from videos.
\newblock In {\em Proc. of International Conference on Machine Learning}, 2022.

\bibitem{xiao2022masked}
Tete Xiao, Ilija Radosavovic, Trevor Darrell, and Jitendra Malik.
\newblock Masked visual pre-training for motor control.
\newblock {\em arXiv preprint arXiv:2203.06173}, 2022.

\bibitem{li2024auxiliary}
Siyuan Li, Shijie Han, Yingnan Zhao, By~Liang, and Peng Liu.
\newblock Auxiliary reward generation with transition distance representation learning.
\newblock {\em arXiv preprint arXiv:2402.07412}, 2024.

\bibitem{luo2024pre}
Hao Luo, Bohan Zhou, and Zongqing Lu.
\newblock Pre-trained visual dynamics representations for efficient policy learning.
\newblock In {\em Proc. of European Conference on Computer Vision}, 2024.

\bibitem{baker2022video}
Bowen Baker, Ilge Akkaya, Peter Zhokov, Joost Huizinga, Jie Tang, Adrien Ecoffet, Brandon Houghton, Raul Sampedro, and Jeff Clune.
\newblock Video pretraining (vpt): Learning to act by watching unlabeled online videos.
\newblock In {\em Proc. of Neural Information Processing Systems}, 2022.

\bibitem{schmeckpeper2021reinforcement}
Karl Schmeckpeper, Oleh Rybkin, Kostas Daniilidis, Sergey Levine, and Chelsea Finn.
\newblock Reinforcement learning with videos: Combining offline observations with interaction.
\newblock In {\em Proc. of Conference on Robot Learning}, 2021.

\bibitem{ye2022become}
Weirui Ye, Yunsheng Zhang, Pieter Abbeel, and Yang Gao.
\newblock Become a proficient player with limited data through watching pure videos.
\newblock In {\em International Journal of Computer Vision}, 2022.

\bibitem{kareer2024egomimic}
Simar Kareer, Dhruv Patel, Ryan Punamiya, Pranay Mathur, Shuo Cheng, Chen Wang, Judy Hoffman, and Danfei Xu.
\newblock {EgoMimic}: Scaling imitation learning via egocentric video.
\newblock {\em arXiv preprint arXiv:2410.24221}, 2024.

\bibitem{singh2024hand}
Himanshu~Gaurav Singh, Antonio Loquercio, Carmelo Sferrazza, Jiajun Wu, Haozhi Qi, Pieter Abbeel, and Jitendra Malik.
\newblock Hand-object interaction pretraining from videos.
\newblock In {\em Proc. of International Conference on Robotics and Automation}, 2025.

\bibitem{luo2025learning}
Hao Luo and Zongqing Lu.
\newblock Learning video-conditioned policy on unlabelled data with joint embedding predictive transformer.
\newblock In {\em International Journal of Computer Vision}, 2025.

\bibitem{zheng2025flare}
Ruijie Zheng, Jing Wang, Scott Reed, Johan Bjorck, Yu~Fang, Fengyuan Hu, Joel Jang, Kaushil Kundalia, Zongyu Lin, Loic Magne, et~al.
\newblock Flare: Robot learning with implicit world modeling.
\newblock {\em arXiv preprint arXiv:2505.15659}, 2025.

\bibitem{romero2017mano}
Javier Romero, Dimitrios Tzionas, and Michael~J Black.
\newblock Embodied hands: Modeling and capturing hands and bodies together.
\newblock {\em ACM Transactions on Graphics}, 36(6):245:1--245:17, 2017.

\bibitem{yang2023hifi}
Dongchao Yang, Songxiang Liu, Rongjie Huang, Jinchuan Tian, Chao Weng, and Yuexian Zou.
\newblock Hifi-codec: Group-residual vector quantization for high fidelity audio codec.
\newblock {\em arXiv preprint arXiv:2305.02765}, 2023.

\bibitem{jaegle2021perceiver}
Andrew Jaegle, Felix Gimeno, Andy Brock, Oriol Vinyals, Andrew Zisserman, and Joao Carreira.
\newblock Perceiver: General perception with iterative attention.
\newblock In {\em Proc. of International Conference on Machine Learning}, 2021.

\bibitem{shridhar2023perceiver}
Mohit Shridhar, Lucas Manuelli, and Dieter Fox.
\newblock Perceiver-actor: A multi-task transformer for robotic manipulation.
\newblock In {\em Proc. of Conference on Robot Learning}, 2023.

\bibitem{simeoni2025dinov3}
Oriane Sim{\'e}oni, Huy~V Vo, Maximilian Seitzer, Federico Baldassarre, Maxime Oquab, Cijo Jose, Vasil Khalidov, Marc Szafraniec, Seungeun Yi, Micha{\"e}l Ramamonjisoa, et~al.
\newblock Dinov3.
\newblock {\em arXiv preprint arXiv:2508.10104}, 2025.

\bibitem{bardes2024vjepa}
Adrien Bardes, Quentin Garrido, Jean Ponce, Xinlei Chen, Michael Rabbat, Yann LeCun, Mido Assran, and Nicolas Ballas.
\newblock V-{JEPA}: Latent video prediction for visual representation learning, 2024.

\bibitem{zhang2025hawor}
Jinglei Zhang, Jiankang Deng, Chao Ma, and Rolandos~Alexandros Potamias.
\newblock Hawor: World-space hand motion reconstruction from egocentric videos.
\newblock In {\em Proc. of Computer Vision and Pattern Recognition}, 2025.

\bibitem{potamias2025wilor}
Rolandos~Alexandros Potamias, Jinglei Zhang, Jiankang Deng, and Stefanos Zafeiriou.
\newblock Wilor: End-to-end 3d hand localization and reconstruction in-the-wild.
\newblock In {\em Proceedings of the Computer Vision and Pattern Recognition Conference}, pages 12242--12254, 2025.

\bibitem{chen2024internvl}
Zhe Chen, Jiannan Wu, Wenhai Wang, Weijie Su, Guo Chen, Sen Xing, Muyan Zhong, Qinglong Zhang, Xizhou Zhu, Lewei Lu, et~al.
\newblock {InternVL}: Scaling up vision foundation models and aligning for generic visual-linguistic tasks.
\newblock {\em arXiv preprint arXiv:2312.14238}, 2024.

\bibitem{lee2025molmoact}
Jason Lee, Jiafei Duan, Haoquan Fang, Yuquan Deng, Shuo Liu, Boyang Li, Bohan Fang, Jieyu Zhang, Yi~Ru Wang, Sangho Lee, et~al.
\newblock Molmoact: Action reasoning models that can reason in space.
\newblock {\em arXiv preprint arXiv:2508.07917}, 2025.

\bibitem{walke2023bridgedata}
Homer~Rich Walke, Kevin Black, Tony~Z Zhao, Quan Vuong, Chongyi Zheng, Philippe Hansen-Estruch, Andre~Wang He, Vivek Myers, Moo~Jin Kim, Max Du, et~al.
\newblock Bridgedata v2: A dataset for robot learning at scale.
\newblock In {\em Conference on Robot Learning}, pages 1723--1736. PMLR, 2023.

\bibitem{liu2025trivla}
Zhenyang Liu, Yongchong Gu, Sixiao Zheng, Xiangyang Xue, and Yanwei Fu.
\newblock Trivla: A unified triple-system-based unified vision-language-action model for general robot control.
\newblock {\em arXiv preprint arXiv:2507.01424}, 2025.

\bibitem{zhang20254d}
Jiahui Zhang, Yurui Chen, Yueming Xu, Ze~Huang, Yanpeng Zhou, Yu-Jie Yuan, Xinyue Cai, Guowei Huang, Xingyue Quan, Hang Xu, et~al.
\newblock 4d-vla: Spatiotemporal vision-language-action pretraining with cross-scene calibration.
\newblock {\em arXiv preprint arXiv:2506.22242}, 2025.

\bibitem{chi2023diffusion}
Cheng Chi, Zhenjia Xu, Siyuan Feng, Eric Cousineau, Yilun Du, Benjamin Burchfiel, Russ Tedrake, and Shuran Song.
\newblock Diffusion policy: Visuomotor policy learning via action diffusion.
\newblock {\em The International Journal of Robotics Research}, 2023.

\bibitem{zhang2025uniact}
Jinliang Zheng, Jianxiong Li, Dongxiu Liu, Yinan Zheng, Zhihao Wang, Zhonghong Ou, Yu~Liu, Jingjing Liu, Ya-Qin Zhang, and Xianyuan Zhan.
\newblock Universal actions for enhanced embodied foundation models.
\newblock In {\em Proc. of Computer Vision and Pattern Recognition}, 2025.

\bibitem{qu2025spatialvla}
Delin Qu, Haoming Song, Qizhi Chen, Yuanqi Yao, Xinyi Ye, Yan Ding, Zhigang Wang, JiaYuan Gu, Bin Zhao, Dong Wang, et~al.
\newblock Spatialvla: Exploring spatial representations for visual-language-action model.
\newblock {\em arXiv preprint arXiv:2501.15830}, 2025.

\bibitem{hou2024diffusion}
Zhi Hou, Tianyi Zhang, Yuwen Xiong, Hengjun Pu, Chengyang Zhao, Ronglei Tong, Yu~Qiao, Jifeng Dai, and Yuntao Chen.
\newblock Diffusion transformer policy.
\newblock {\em arXiv preprint arXiv:2410.15959}, 2024.

\bibitem{huang2025thinkact}
Chi-Pin Huang, Yueh-Hua Wu, Min-Hung Chen, Yu-Chiang~Frank Wang, and Fu-En Yang.
\newblock Thinkact: Vision-language-action reasoning via reinforced visual latent planning.
\newblock {\em arXiv preprint arXiv:2507.16815}, 2025.

\end{thebibliography}
